\documentclass{article}

\usepackage[main, final]{neurips_2026}

\usepackage[utf8]{inputenc} 
\usepackage[T1]{fontenc}    
\usepackage{hyperref}       
\usepackage{url}            
\usepackage{amsfonts}       
\usepackage{nicefrac}       
\usepackage{microtype}      
\usepackage{graphicx}
\usepackage{soul}
\usepackage{diagbox}
\usepackage{amsmath}
\usepackage{amssymb}
\usepackage{mathtools}
\usepackage{amsthm}
\usepackage{multirow}
\usepackage{enumitem}
\usepackage{subcaption}
\usepackage{booktabs}    
\usepackage[table]{xcolor} 
\usepackage{colortbl} 
\usepackage{pifont}
\usepackage{makecell}
\usepackage{tabularx}

\newcolumntype{C}{>{\centering\arraybackslash}X}
\definecolor{chead}{RGB}{220,225,240}        
\definecolor{cbin}{RGB}{205,232,205}          
\definecolor{cbinfade}{RGB}{232,242,232}      
\definecolor{cmarkgreen}{RGB}{50,130,60}      
\definecolor{cspdL}{RGB}{255,247,222}         
\definecolor{cspdM}{RGB}{255,224,150}         
\definecolor{cspdH}{RGB}{245,180,30}           
\definecolor{cbdL}{RGB}{248,225,225}          
\definecolor{cbdM}{RGB}{232,150,110}          
\definecolor{cbdH}{RGB}{200,90,55}              
\definecolor{ctradband}{RGB}{180,180,180}     
\definecolor{cganband}{RGB}{170,140,40}        
\definecolor{cdiffband}{RGB}{70,110,180}       
\definecolor{coursband}{RGB}{210,90,40}         
\definecolor{ctradrow}{RGB}{240,240,240}      
\definecolor{cganrow}{RGB}{255,247,222}        
\definecolor{cdiffrow}{RGB}{225,235,250}      
\definecolor{coursrow}{RGB}{255,228,210}      
\usepackage[table]{xcolor}
\definecolor{Group1Color}{HTML}{B3D9FF}  
\definecolor{Group2Color}{HTML}{B3FFB3}  
\definecolor{Group3Color}{HTML}{FFE699}  
\definecolor{Group4Color}{HTML}{FFC2E0}  
\definecolor{Group5Color}{HTML}{D9C2FF}  

\definecolor{Group6Color}{HTML}{5FC2B8}  
\definecolor{Group7Color}{HTML}{D2A86E}  
\definecolor{Group8Color}{HTML}{8A9BB3}  

\newcommand{\cmark}{\ding{51}}

\title{PerQ: Inverse Generative Modeling for Neural Image Compression via Quantization Error Compensation}

\author{%
\textbf{Tianhao Peng}$^{1}$,
\textbf{Ho~Man~Kwan}$^{1}$,
\textbf{Fan~Zhang}$^{1}$, 
\textbf{Shan Liu}$^{2}$,
\textbf{Dave~Bull}$^{1}$ \\
$^{1}$Visual Information Lab, University of Bristol, UK \\
$^{2}$Tencent Media Lab, Palo Alto, USA \\
\texttt{\{tianhao.peng, hm.kwan, fan.zhang, dave.bull\}@bristol.ac.uk} \\
\texttt{shanl@global.tencent.com}
}
\begin{document}

\maketitle

\begin{abstract}
Neural image codecs face a fundamental trade-off between pixel fidelity and perceptual quality. Distortion-optimized codecs preserve pixel-wise fidelity but produce blurred reconstructions at low bitrates, whereas generative codecs leverage learned priors to improve perceptual quality but tend to exhibit saturating PSNR at higher bitrates and require separate models for different bitrate targets. We propose \textbf{PerQ}, which reframes perceptual reconstruction as inverse generative modeling with bounded support. Specifically, we define the quantization cell as the set of latents that yield the same transmitted symbols and derive a transport bound showing that reconstructions from this cell remain within a rate-determined radius of the codec output. This motivates performing generative reconstruction within the codec-induced quantization cell, rather than sampling from an unconstrained image prior. PerQ implements this principle by attaching a rate-conditioned flow-matching compensator to a frozen, multi-rate distortion-optimized backbone, enabling perceptual enhancement while preserving the backbone's coding performance. Experiments on Kodak and CLIC2020 show that PerQ achieves perceptual performance comparable to generative codecs such as MS-ILLM in LPIPS and FID, while maintaining competitive PSNR relative to distortion-optimized codecs such as ELIC. A single model checkpoint supports a wide bitrate range with fast decoding, and the compensator provides consistent gains across three different backbones. The implementation is available at \url{https://jasminepp.github.io/PerQ/}.

\end{abstract}
\vspace{-5pt}
\section{Introduction}
\label{sec:intro}

Image compression has long been a fundamental research topic in signal processing and a core component of modern visual data infrastructure \cite{bull2021intelligent}. Conventional image codecs such as JPEG \cite{wallace1991jpeg}, BPG \cite{bellard2017bpg}, and the intra modes of AV1 \cite{av1} and VVC \cite{bross2021overview}, have been widely adopted over the past decades. More recently, end-to-end learned image compression \cite{balle2016end, balle2018variational,minnen2020channel} have achieved competitive or superior rate-distortion performance compared with conventional codecs based on pixel-wise metrics, such as PSNR \cite{wang2009mean}. However, optimizing pixel-wise distortion does not necessarily yield perceptually pleasing reconstructions, especially at low bitrates, where severe bitrate constraints limit the preservation of fine details and high-frequency image content, often resulting in visually blurred reconstructions. This limitation has motivated growing interest in perceptual image compression, which prioritizes perceptual fidelity over pixel-wise similarity and has led to the development of perceptual and generative image codecs.

Recent generative codecs  have demonstrated promising perceptual quality at extremely low bitrates \cite{jia2025cod,zhang2025ultra}. However, extending this perceptual advantage to practical deployment remains challenging. (i) \emph{Pixel-wise fidelity}: downstream applications often require reconstructions to remain faithful to the source pixels, but most generative codecs, whose reconstructions are dominated by a generative prior, tend to plateau on PSNR even as the bitrate becomes high, since additional bits do not bring the prior closer to the source pixels \cite{mentzer2020high, muckley2023improving}. (ii) \emph{Bitrate coverage}: practical deployment would benefit from a single model that supports a broad range of bitrates, avoiding the need to store and maintain separate models for different rates. However, most existing perceptual codecs require separate models for different target rates \cite{mentzer2020high,zhang2025stablecodec}. Although a few approaches support multiple bitrates with a single model \cite{careil2023towards,guo2025oscar}, their operating ranges are typically concentrated in a relatively narrow low-bitrate window. 
The results of existing works shown in \autoref{tab:positioning} reveal that none of them delivers high perceptual quality, competitive pixel fidelity, and wide bitrate coverage with a single trained model.
\begin{figure}
    \centering
    \includegraphics[width=1\linewidth]{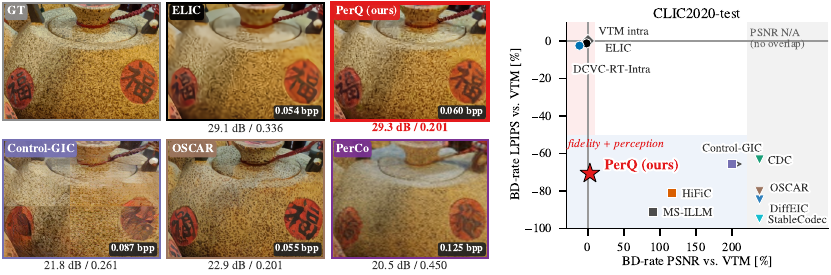}
    \caption{\textbf{Balance between Perceptual quality and pixel fidelity.}
(Left) Visual comparison at comparable bitrates, with PSNR (dB) / LPIPS. 
(Right) BD-rate of PSNR against BD-rate of LPIPS on CLIC2020-test, both relative to VTM intra, lower is better).}
\label{fig:teaser}
\vspace{-15pt}
\end{figure}

 We ask, \textit{Can these competing objectives be reconciled within a single codec without sacrificing the advantages of distortion-optimized compression?} We propose \textbf{PerQ}, an image codec that reframes perceptual reconstruction as \emph{inverse generative modeling with bounded support}, leveraging the quantization structure of an existing codec to constrain perceptual refinement. We characterize the set of latent representations that yield the same transmitted symbols as a \emph{quantization cell} and derive a \emph{transport bound} relating the quantization step to a rate-determined radius in pixel space. Unlike prior approaches that augment distortion-optimized backbones with perceptual losses~\cite{he2022po, korber2024egic, mentzer2020high} and use learned generative priors to explore the image manifold, PerQ employs a flow-matching-based compensator as a decoder-side plug-in to improve perceptual quality while keeping the refined latent within this bounded region. Integrated into the decoder, the compensator leaves the bitstream unchanged and supports the backbone's existing multi-rate operating range with a single model, while retaining fast decoding. Importantly, the derived transport bound guarantees that the refined reconstruction remains within a rate-determined pixel-space radius of the backbone's output, even with perceptual optimization. This design brings together the perceptual advantages of generative codecs, the pixel fidelity of distortion-optimized codecs, and broad bitrate coverage within a single model (\autoref{fig:teaser}). The main contributions of this work are:
\vspace{-5pt}
\begin{itemize}[leftmargin=*]
   \item \textbf{(C1) Inverse generative modeling with bounded support.} We formulate perceptual reconstruction as inverse generative modeling over a rate-determined bounded support, characterized by a transport bound (\S\ref{sec:method:transport}) that relates the quantization step to a pixel-space radius around the output of a multi-rate distortion-optimized codec.

   \item \textbf{(C2) A rate-conditioned flow-matching compensator design} is designed (\S\ref{sec:method:design}) for perceptual reconstruction within the bounded support. A single checkpoint supports multiple backbone rates, while the decoder-side design preserves the bitstream and generalizes across different backbones.

   \item \textbf{(C3) PerQ: A perceptual-fidelity-balanced image codec} integrates a bounded-support compensator with a frozen distortion-optimized backbone, bridging the gap between generative perceptual codecs and distortion-oriented codecs by combining strong perceptual quality with competitive pixel fidelity in a single, multi-rate model.
\end{itemize}
\vspace{-5pt}
Extensive experiments on Kodak and CLIC2020-test (\autoref{tab:positioning}) show that PerQ retains most of the perceptual gains of strong generative codecs such as MS-ILLM while achieving competitive pixel fidelity against VVC (VTM intra), while enabling efficient four-step decoding at $86$ ms/MP. Moreover, a single checkpoint covers nearly two orders of magnitude in bitrate, and the same compensator delivers consistent gains across two additional backbones. These results demonstrate that PerQ balances perceptual quality, pixel fidelity, and bitrate flexibility within a single, efficient codec.

\begin{table}[t]
\caption{\textbf{Positioning of image codecs across
compression paradigms.} Each row reports single-model multi-rate operation, decoding complexity and BD-rate against VTM intra
on the Kodak~\cite{kodak1993} and CLIC2020-test~\cite{toderici2020clic} datasets.}

\resizebox{\linewidth}{!}{\begin{tabular}{rccccccccccc}
\toprule
 & & & & \multicolumn{4}{c}{\textbf{Kodak}: BD-rate vs.\ VVC (\%) $\downarrow$}
 & \multicolumn{4}{c}{\textbf{CLIC2020-test}: BD-rate vs.\ VVC (\%) $\downarrow$} \\
\cmidrule(lr){5-8} \cmidrule(lr){9-12}
\multirow{-2}{*}{\textbf{Image codecs}}
 & \multirow{-2}{*}{\shortstack[c]{Single-model\\ multi-rate}}
 & \multirow{-2}{*}{\shortstack[c]{Dec.\ time\\ (ms/Mp) $\downarrow$}}
 & \multirow{-2}{*}{\shortstack[c]{Dec.\\ kMACs/px $\downarrow$}}
 & PSNR & LPIPS & DISTS & FID & PSNR & LPIPS & DISTS & FID \\
\midrule
\multicolumn{12}{l}{\textbf{\color{ctradband}Conventional codec}} \\
    VTM intra~\cite{bross2021overview} & -- & -- & -- & 0 & 0 & 0 & 0 & 0 & 0 & 0 & 0 \\
\midrule
\multicolumn{12}{l}{\textbf{\color{ctradband}VAE-based codecs}} \\
    ELIC~\cite{he2022elic} &  & 206 & 332 & $-2.0$ & $-2.1$ & $+2.4$ & $+7.3$ & $-1.9$ & $-1.3$ & $+9.4$ & $-13.8$ \\
    EVC~\cite{wang2023evc} & \cmark & 57 & 291 & $+2.0$ & $-9.0$ & $-5.4$ & $-4.0$ & $+3.7$ & $-8.7$ & $-2.3$ & $-6.2$ \\
    DCVC-RT-Intra~\cite{jia2025towards} & \cmark & 46 & 362 & $-11.2$ & $-8.7$ & $-2.3$ & $-1.8$ & $-11.8$ & $-2.5$ & $+8.8$ & $-30.1$ \\
\midrule
\multicolumn{12}{l}{\textbf{\color{cganband}GAN-based perceptual codecs}} \\
    MS-ILLM~\cite{muckley2023improving} &  & 150 & 675 & $+85.9$ & $-86.4$ & $-82.9$ & $-86.3$ & $+89.9$ & $-91.0$ & $-90.7$ & $-97.4$ \\
    HiFiC~\cite{mentzer2020high} &  & 150 & 546 & $+128.6$ & $-69.8$ & $-59.0$ & $-64.3$ & $+116.3$ & $-81.1$ & $-76.9$ & $-83.0$ \\
    Control-GIC~\cite{li2024once} & \cmark & 416 & 6299 & $+372.4$ & $-74.0$ & $-68.5$ & $-69.7$ & $+1045.8$ & $-65.6$ & $-73.8$ & $-85.2$ \\
\midrule
\multicolumn{12}{l}{\textbf{\color{cdiffband}Diffusion-based perceptual codecs}} \\
    PerCo~\cite{careil2023towards} &  & 6822 (20-step) & 30930 & N/A & $-64.1$ & $-84.4$ & $-92.9$ & N/A & N/A & N/A & $-94.3$ \\
    DiffEIC~\cite{li2024toward} &  & 11628 (50-step) & 89899 & N/A & $-81.0$ & $-88.8$ & $-91.3$ & N/A & $-84.6$ & $-87.4$ & $-96.5$ \\
    CDC~\cite{yang2023lossy} &  & 4854 (50-step) & 39440 & $+318.7$ & $-48.6$ & $-71.7$ & $-68.8$ & N/A & $-63.3$ & $-74.7$ & $-86.8$ \\
    StableCodec~\cite{zhang2025stablecodec} &  & 494 (1-step) & 6138 & N/A & $-92.7$ & $-96.2$ & $-96.8$ & N/A & $-94.9$ & $-97.1$ & $-99.0$ \\
    OSCAR~\cite{guo2025oscar} & \cmark & 456 (1-step) & 8262 & N/A & $-79.0$ & $-91.8$ & $-92.3$ & N/A & $-80.2$ & $-88.7$ & $-96.0$ \\
\midrule
    \textbf{PerQ (Ours)} & \cmark & \textbf{86 (4-step)} & \textbf{702} & {\boldmath$-1.6$} & {\boldmath$-57.4$} & {\boldmath$-63.2$} & {\boldmath$-67.9$} & {\boldmath$+2.7$} & {\boldmath$-70.6$} & {\boldmath$-77.4$} & {\boldmath$-91.9$} \\
\bottomrule
\end{tabular}}
{\flushleft \fontsize{6pt}{7pt}\selectfont\color{black!55}
$^{\ast}$ All BD-rates anchored to VVC intra (VTM 22.0, AI mode, YCbCr 4:4:4 BT.709, 10-bit)~\cite{bross2021overview}, sweeping six QP values (17, 22, 27, 32, 37, 42); N/A: the quality ranges do not overlap. $^{\dagger}$ For neural codecs that ship multiple checkpoints, we report decoding time and complexity on the mid-rate checkpoint.}
\label{tab:positioning}
\vspace{-10pt}
\end{table}
\section{Related Work}
\label{sec:related}

\noindent\textbf{Learned image compression} originates from the joint optimization of an analysis/synthesis transform and a
learned entropy model \cite{theis2017lossy, balle2016end}, with subsequent progress driven primarily by more effective entropy models and stronger transforms. Hyperpriors capture spatially-varying scale \cite{balle2018variational}; discretized Gaussian mixture likelihoods refine the conditional entropy
model \cite{cheng2020learned}; channel-wise autoregressive factorizations sharpen it further \cite{minnen2020channel}; checkerboard contexts parallelize spatial autoregression \cite{he2021checkerboard}; and recent designs combine grouped channel attention \cite{he2022elic} with multi-reference context \cite{jiang2023mlic, jiang2025mlic++} or transformer-CNN hybrid transforms \cite{liu2023learned}. The best performing neural image codecs match or exceed VVC all intra mode on both PSNR and MS-SSIM, with real-time variants closing the latency gap \cite{jia2025towards}, and they remain visually faithful across moderate to high bitrates. The picture changes at low rates, where the bit budget is limited to preserve high-frequency content, and reconstructions become visibly blurred. The results in \cite{blau2018perception,blau2019rethinking} show that the divergence between source and reconstruction marginals widens at low rates and cannot be closed with further distortion-oriented training.

\noindent\textbf{Generative image codecs} are designed to primarily target lower bitrates by conditioning reconstructions on a generative prior trained to match natural-image statistics, trading pixel-wise fidelity for perceptual realism. GAN-based codecs integrate adversarial training into the rate-distortion objective \cite{tschannen2018deep, agustsson2019generative,mentzer2020high, muckley2023improving, he2022po}, with subsequent work exposing the distortion-perception trade-off during decoding \cite{agustsson2023multi,korber2024egic} and obtaining variable-rate operation from a single codec \cite{li2024once}. Alternatively, diffusion-based codecs employ learned denoising at the decoder, with early work casting compression as reverse channel coding of noise samples through a Gaussian diffusion process \cite{theis2022lossy}. Follow-up studies generalize this to pretrained foundation priors \cite{vonderfecht2025lossy} and perform multi-rate operations on top of a frozen prior \cite{jia2025cod}. A separate line trains a conditional diffusion model as the codec decoder. CDC \cite{yang2023lossy} replaces the VAE decoder with a conditional diffusion model, PerCo \cite{careil2023towards} couples a spatial latent with a global textual code for ultra-low-rate decoding, and other designs guide a frozen text-to-image or semantic prior \cite{lei2023text+, li2024toward,relic2024lossy,hoogeboom2023high} with compressed content. These designs typically incur multi-step sampling costs during decoding. More recent work distills the iterative process into a single
step \cite{jia2024generative, zhang2025stablecodec,guo2025oscar, xue2025one, zhang2025ultra}. Since reconstruction is conditioned on the generative prior rather than driven by the bitstream alone, pixel-space fidelity saturates as bpp increases, and PSNR improves sub-linearly with rate. Moreover, most designs commit to a single rate point at training time, requiring a separate model per operating point on the rate-distortion curve.


\section{Method}

\label{sec:method}

As mentioned in Section \ref{sec:intro}, unconstrained generative compression typically conditions reconstructions on a learned image prior, often incurring a penalty in pixel-wise fidelity as a result. This design is acceptable in applications that can tolerate hallucinated details and deviations from source fidelity, as opposed to distortion-oriented codecs. In this paper, we aim to close the perceptual gap of distortion-optimized image codecs while preserving reconstructions that remain faithful to the source. To this end, we constrain
perceptual reconstruction within the quantization-error space induced by the compression backbone. In this section, we first derive the structural properties of this constraint in \S\ref{sec:method:transport} and \S\ref{sec:method:design:support}. We then propose a flow-matching based compensator in \S\ref{sec:method:design}, which is integrated with a compression codec backbone (\S\ref{sec:method:backbone}). We describe the training methodology in \S\ref{sec:method:training}.

\subsection{The Transport Bound}
\label{sec:method:transport}
Let $x \in \mathbb{R}^{H \times W \times 3}$ denote the source image, $f^{\mathrm{enc}}_{\phi}$ the encoder, $f^{\mathrm{dec}}_{\theta}$ the decoder, and $y = f^{\mathrm{enc}}_{\phi}(x) \in \mathbb{R}^{h \times w \times C}$ the latent. In a conventional setting, quantization is performed using a positive scale factor $q$, which maps the latent into the symbol space (i.e., the integer domain) for rounding. The complete quantization and decoding process is given by:
\begin{align}
\text{Quantization:}& \quad \hat{y}^{\mathrm{sym}} = \lfloor y^{\mathrm{sym}} \rceil = \lfloor \frac{y}{q} \rceil, \\
\text{Decoding:}& \quad \hat{x} = f^{\mathrm{dec}}_{\theta}(\hat{y}), \quad \text{where} \quad \hat{y} = q \cdot \hat{y}^{\mathrm{sym}}.
\label{eq:notation}
\end{align}
in which $q \in \mathbb{R}_+^{h, w, C}$ is the quantization
step size, which may vary over spatial locations and channels
in modern neural compression
models \cite{wang2023evc, jia2025towards}. Following the common
practice \cite{balle2018variational}, the
non-differentiable rounding operation is emulated during
training with additive uniform noise \cite{balle2016end} or a
straight-through estimator
(STE) \cite{theis2017lossy, bengio2013estimating}. In this paper, we utilize the latter in the reconstruction path, together with a distortion loss.

Under round-to-nearest quantization, the latent error in the symbol space is bounded by $\|y^{\mathrm{sym}} - \hat y^{\mathrm{sym}}\|_\infty \le \tfrac{1}{2}$ \cite{gersho2012vector}. Rescaling by $q$ gives the following element-wise bound in the original latent space:
\begin{equation}
  | y_{i, j, c} - \hat{y}_{i, j, c} | \le
  \tfrac{q_{i, j, c}}{2},
  \quad \text{for } i, j, c,
  \label{eq:latent_bound}
\end{equation}
where $i, j, c$ are the spatial coordinates and channel index of the latent.

To translate the latent-space quantization bound into a bound on reconstruction deviation, we exploit the local Lipschitz continuity of the decoder $f^{\mathrm{dec}}_\theta$ over a compact operating range \cite{virmaux2018lipschitz}. By \eqref{eq:latent_bound}, the source latent $y$ lies in the \textbf{\emph{quantization cell}} of $\hat y$, defined as the set of latents within $q/2$ of $\hat y$ element-wise. For each latent channel $c$, the Lipschitz constant of the decoder with respect to that channel,
\begin{equation}
  L_c \;:=\; \sup_{u \in \mathcal{Y}} \left\| \tfrac{\partial f^{\mathrm{dec}}_\theta(u)}{\partial u_{:,:,c}} \right\|_{\mathrm{op}},
  \label{eq:Lc_def}
\end{equation}
measures the sensitivity of the full output to a perturbation confined to channel $c$ of the input, where the partial derivative is taken over the spatial slice of channel $c$ at the points where it exists, and $\mathcal{Y}$ is an operating range that contains every quantization cell of the backbone, so that the segment between $\hat y$ and any latent of its cell lies in $\mathcal{Y}$. By \eqref{eq:latent_bound}, each per-channel difference $\| y_{: ,:,c} - \hat y_{:,:,c} \|_2$ is at most $\tfrac{q_c \sqrt{hw}}{2}$, where $q_c := \max_{i,j} q_{i,j,c}$ is the largest step of channel $c$. Composing with \eqref{eq:Lc_def} and summing over channels yields the \emph{transport bound}:
\begin{equation}
  \| f^{\mathrm{dec}}_\theta(y) - f^{\mathrm{dec}}_\theta(\hat y) \|_2 \;\le\; \sum_{c=1}^{C} L_c \, \| y_{:,:,c} - \hat y_{:,:,c} \|_2 \;\le\; \tfrac{\sqrt{hw}}{2} \sum_{c=1}^{C} L_c \, q_c \;=:\; R(q).
  \label{eq:transport}
\end{equation}
The transport bound establishes that every reconstruction decoded from the quantization cell, including the source latent $y$, lies within a radius $R(q)$ of the backbone reconstruction. Importantly, this radius is linear in the quantization step, with the decoder contributing only the fixed constants $L_c$. Although the bound is loose due to the supremum-based sensitivity estimates and the nonuniform distribution of actual quantization errors (\autoref{fig:latent_error_dist}), its key value lies in characterizing a rate-dependent bounded region for reconstruction rather than providing a tight numerical estimate.

\subsection{Bounded-support Modeling}
\label{sec:method:design:support}
The transport bound turns the quantization cell from an error characterization into a feasible region for perceptual reconstruction. In the latent space, \eqref{eq:latent_bound} defines the quantization cell $\mathcal{C}(\hat{y}, q) = \{\, y' :\|y' - \hat{y}\|_{\infty} \le q/2 \,\}$ around the latent $\hat y$: the source latent $y$ lies in this cell, and every latent in it rounds to the same transmitted symbols. In the pixel space, \eqref{eq:transport} confines every reconstruction decoded from the cell to the closed ball $\mathcal{B}(\hat x, q) := \{\, u : \|u - \hat x\|_2 \le R(q) \,\}$ around the backbone reconstruction $\hat x = f^{\mathrm{dec}}_\theta(\hat y)$. This ball constrains decoder outputs relative to $\hat x$, rather than bounding the reconstruction error relative to the source image $x$, which also depends on the distortion introduced by the encoder-decoder transform. This transform-induced distortion remains fixed once the backbone is frozen (\S\ref{sec:method:backbone}).

The rate-distortion-perception trade-off \cite{blau2018perception, blau2019rethinking} implies that a distortion-optimized backbone may retain a perceptual distribution gap $d_p(p_X,p_{\hat X})>0$. Rather than recovering the source latent itself, we formulate perceptual reconstruction as inverse generative modeling over the quantization cell, treating it as a feasible design space in which to seek perceptually enhanced reconstructions. Accordingly, we introduce a compensator \(\psi\) that models the conditional distribution:
\begin{equation}
\tilde{y} \sim p_{\psi}(\tilde{y} \mid \hat{y}, \hat{z}, q),
\qquad
\mathrm{supp}\,p_{\psi} \subseteq \mathcal{C}(\hat{y}, q),
\label{eq:cell_constraint}
\end{equation}
in which $\hat z$ is the auxiliary latent produced by the backbone's hyper-prior and the support is enforced by clamping at the decoder side (\S\ref{sec:compensator_arch}). Every sample $\tilde y$ quantizes to the same transmitted symbols, and its decoded image lies within $\mathcal{B}(\hat x, q)$. Unlike prior work \cite{yi2026rethinking}, which models quantization error as a latent-recovery problem by adding learned noise to $\hat y$ to approximate the source latent $y$, PerQ treats the quantization cell as a design space. Any $\tilde y$ within the cell is a valid reconstruction candidate, and the goal is to find one whose decoded image better closes the perceptual gap, rather than to recover $y$ exactly. Moreover, as shown in \autoref{app:bounded_support}, matching the conditional distribution of the source latent would reproduce the distribution of the unquantized reconstruction.

\subsection{Overall framework}
\label{sec:method:design}

\begin{figure}[t]
    \centering
    \includegraphics[width=1\linewidth]{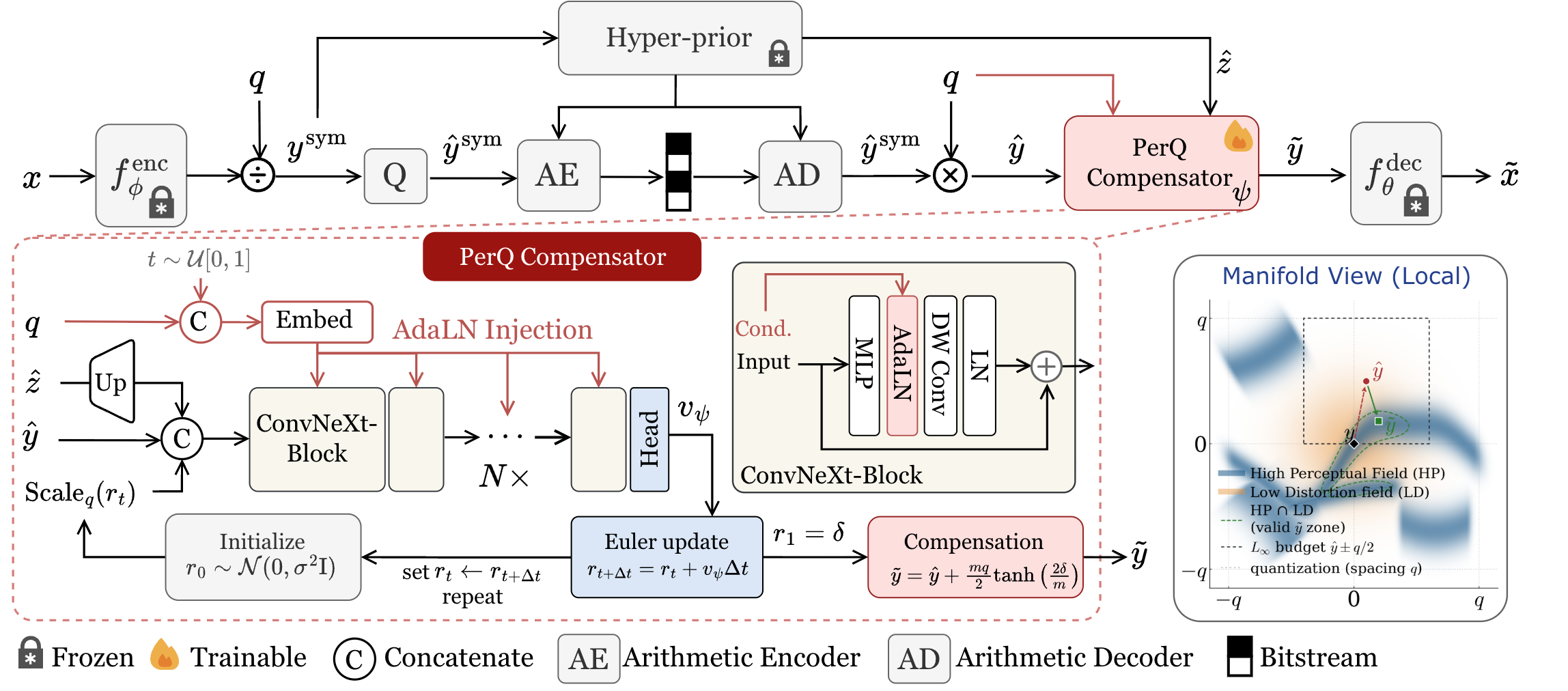}
    \caption{\textbf{PerQ overview.} \textbf{Top:} the
compression pipeline. 
\textbf{Bottom-left:} architecture of the flow-matching based compensator.
\textbf{Bottom-right:} illustrative 2D slice of latent
space. The cell is centered at the quantized
latent $\hat y$ and contains the source latent $y$ by the quantization bound \eqref{eq:latent_bound}. The compensator generates
$\tilde y$ within the cell, targeting the
high-perceptual-quality field (\textcolor{blue}{blue}) while remaining inside the low-distortion field (\textcolor{orange}{orange}).}
    \label{fig:mainframework}
    \vspace{-10pt}
\end{figure}

\autoref{fig:mainframework} provides an overview of the proposed PerQ framework, in which a compensator $\psi$ is proposed to be located between the entropy decoder and the image decoder within the compression backbone. Given the entropy-decoded latent $\hat{y}$ and the auxiliary latent $\hat z$ produced by the backbone's hyper-prior, the compensator generates a refined latent $\tilde{y}$ which satisfies the cell constraint outlined in \eqref{eq:cell_constraint}, which is then passed to the image decoder $f^{\mathrm{dec}}_\theta$ to produce the reconstructed image $\tilde x$.

\subsubsection{Compensator Architecture}
\label{sec:compensator_arch}
\noindent\textbf{Flow-matching parameterization.}
We instantiate the compensator as a flow-matching network \cite{lipman2022flow, liu2022flow}. Specifically, we parameterize a conditional velocity field $v_{\psi}(r_t, t \mid \hat{y}, \hat{z}, q)$ and obtain the symbol-space residual $\delta$ as the final state of an ODE initialized at Gaussian noise: 
\begin{equation}
r_0 \sim \mathcal{N}(0, {\sigma^2}I),
\qquad
\frac{dr_t}{dt} = v_{\psi}(r_t, t \mid \hat{y}, \hat{z}, q),
\qquad
\delta = r_1.
\end{equation}
We solve this ODE using \(K\) Euler steps of size \(\Delta t=1/K\):
\begin{equation}
r_{t + \Delta t} = r_t + v_{\psi}(r_t, t \mid \hat{y}, \hat{z}, q)\,\Delta t,
\qquad t = 0, \tfrac{1}{K}, \dots, \tfrac{K-1}{K}.
\end{equation}

The compensator is trained with the rectified-flow term~\cite{liu2022flow}. We define the training target as the symbol-space quantization error \(\delta_{\text{target}} = y^{\mathrm{sym}} - \hat y^{\mathrm{sym}} = (y - \hat y) / q\), i.e., the displacement from the quantized latent to its pre-rounding source. For each sample, we draw $t \sim \mathcal{U}[0, 1]$ and perform linear interpolation with the noise prior and the target, $r_t = (1 - t) r_0 + t\delta_{\text{target}}$. The velocity field is regressed against the constant velocity along this interpolant:
\begin{equation}
\mathcal{L}_{\text{FM}}
=
\mathbb{E}_{r_0, t, \delta_{\text{target}}}
\bigl\| v_{\psi}(r_t, t \mid \hat{y}, \hat{z}, q) - (\delta_{\text{target}} - r_0)\bigr\|_2^2.
\label{eq:fm_loss}
\end{equation}

We set the noise scale to match the range allowed by the final residual clamp. By default, we set $\sigma = \frac{m}{\sqrt{12}}$, the standard deviation of a uniform distribution over the symbol-space sub-cell $[-m/2, m/2]$ that the bounded compensator can reach, where $m$ is the bound scale (see next). The prior $r_0 \sim \mathcal{N}(0, \sigma^2 I)$ then matches the scale of the compensator's reachable region.

\begin{figure}
    \centering
    \includegraphics[width=0.95\linewidth]{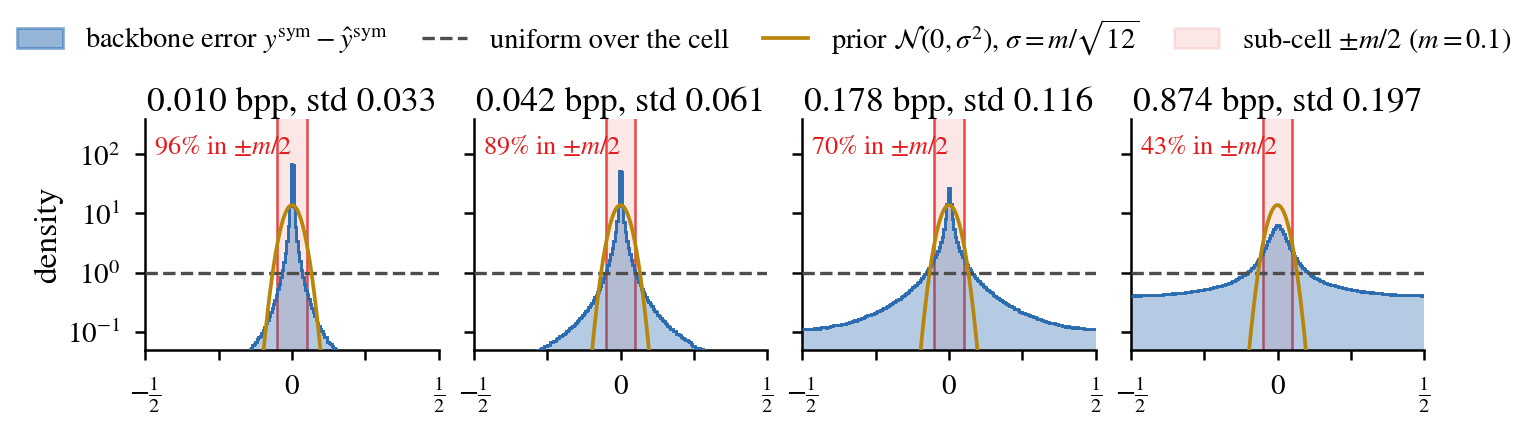}
    \caption{\textbf{Symbol-space distribution of the backbone quantization error} $y^{\mathrm{sym}} - \hat y^{\mathrm{sym}}$ on Kodak at four rate points (log density). The sub-cell $\pm m/2$ (red, $m = 0.1$) contains most of the error at low rates, and the prior $\mathcal{N}(0, \sigma^2)$ with $\sigma = m/\sqrt{12}$ (gold) is set to the scale of this sub-cell.}
\label{fig:latent_error_dist}
\vspace{-10pt}
\end{figure}

\noindent\textbf{Residual clamping.}
The integrated residual $\delta$ is soft-bounded element-wise via a $\tanh$ nonlinearity, and the final compensated latent is
\begin{equation}
 \tilde{y} = \hat{y} + \tfrac{m q}{2}\tanh\left(\tfrac{2\delta}{m}\right),
\label{eq:clamp}
\end{equation}
where $m \in (0, 1]$ is the bound scale that restricts the compensation to $\pm m/2$ in symbol space, a sub-cell of width $m q$ centered at $\hat y$. Since the transport bound \eqref{eq:transport} is linear in the half-width of the cell, a sub-cell of width $m q$ gives a radius $m R(q)$: the decoded reconstruction stays within $m R(q)$ of the backbone reconstruction, so the bound scale is an explicit control of the fidelity cost of perceptual compensation, which the bound-scale sweep of \autoref{tab:ablation_bound} confirms.

Empirically (\autoref{fig:latent_error_dist}), the latent quantization residual occupies a range that varies substantially across rate points, with the bulk of its mass shrinking at low rates. This is because of the variable rate compression model design: at lower rate points, the quantization step $q$ is larger, compacting the latent $y^{\mathrm{sym}} = \frac{y}{q}$ into a small region of symbol space, where the residual inherits this small magnitude. Generating compensations over the full cell would therefore introduce excessive distortion, as a $\pm \tfrac{q}{2}$ perturbation is large relative to the signal. We use $m = 0.1$ by default. While this restricts the compensation magnitude at high rates, distortion-optimized backbones already achieve high quality at high bitrates, so a small-magnitude refinement is sufficient to close the perceptual gap.

\noindent\textbf{Velocity field architecture.} In our implementation, $v_\psi$ is a ConvNeXt network~\cite{liu2022convnet} with $N$ blocks, whose configuration is given in \autoref{app:Implementation}. To keep the velocity field at one operating scale across rate points, we multiply the flow state $r_t$ by $q$ before the network, and divide the output by $q$ to return to symbol space, where the bound of \eqref{eq:clamp} is applied. We observe that this scaling improves performance over operating directly in symbol space. The hyper-prior latent $\hat z$ is upsampled by $4{\times}$ to the spatial resolution of $\hat y$ through two stages of $3{\times}3$ convolution and $2{\times}$ pixel shuffle with a SiLU activation in between. Following DiT~\cite{peebles2023scalable}, the timestep $t$ is embedded by a sinusoidal embedding and an MLP, and enters each block through a zero-initialized linear projection to the shift, scale, and gate of an adaLN-zero modulation.

\subsubsection{Compression backbone}
\label{sec:method:backbone}
We focus on modeling the quantization-error space and adopt an existing neural image compression model as the codec backbone with minor modifications. Specifically, we use the intra-frame model of DCVC-RT \cite{jia2025towards}, which delivers strong rate--distortion performance at low complexity, supports fast encoding and decoding, and provides variable-bitrate coding by selecting a quantization parameter for the target rate point.

One implementation detail of \cite{jia2025towards} is that the quantization step size is applied at two stages on both the encoder and decoder sides. The encoder $f^{\mathrm{enc}}_\phi$ is decomposed into two sub-networks $f^{\mathrm{enc}1}_\phi$ and $f^{\mathrm{enc}2}_\phi$: the rate-specific parameter $q^{\mathrm{enc}}$ is applied between them, and the per-pixel parameter $q^{\mathrm{enc}}_{\mathrm{local}}$ (predicted by the hyper-prior) is applied at the bottleneck:
\begin{equation}
y^{\mathrm{sym}} = f^{\mathrm{enc}2}_\phi\bigl((
  f^{\mathrm{enc}1}_\phi(x) \cdot q^{\mathrm{enc}}\bigr)
  \cdot q^{\mathrm{enc}}_{\mathrm{local}}.
\end{equation}
The decoder mirrors this structure during inverse quantization. While this distributed scaling provides additional
capacity for rate adaptation, it conflicts with the
analysis in \S\ref{sec:method:transport}: when the rate-specific
scaling is absorbed into the encoder/decoder networks,
$f_\phi$ becomes rate-adaptive, so different rate points
produce different bottleneck representations of the same
image, and the encoder--decoder pair itself becomes an information bottleneck. In such a case, a decoder that knows it operates at a low rate can learn to discard detail, e.g., producing blurrier outputs, which empirically lowers the MSE at low rate points.
The quantization error remains bounded in symbol space, but the unrounded decode itself is far from the source at low rates, so the cell no longer contains a high-fidelity latent and in-cell compensation has little headroom.

We therefore modify \cite{jia2025towards} by consolidating all
quantization scaling at the
bottleneck\footnote{For brevity, we omit the quantization
offset (mean subtraction) used in the actual
implementation.}:
\begin{equation}
y^{\mathrm{sym}} = f^{\mathrm{enc}2}_\phi\bigl(
  f^{\mathrm{enc}1}_\phi(x)\bigr)
  \cdot q^{\mathrm{enc}} \cdot q^{\mathrm{enc}}_{\mathrm{local}},
\end{equation}
and symmetrically on the decoder side:
\begin{equation}
\hat{y} = \hat{y}^{\mathrm{sym}}
  \cdot q^{\mathrm{dec}} \cdot q^{\mathrm{dec}}_{\mathrm{local}},
\qquad
\hat{x} = f^{\mathrm{dec}}_\theta(\hat{y})
  = f^{\mathrm{dec}2}_\theta\bigl(f^{\mathrm{dec}1}_\theta(\hat{y})\bigr),
\end{equation}
where the decoder-side scales are paired with the encoder-side ones, $q^{\mathrm{dec}} = 1/q^{\mathrm{enc}}$ and
$q^{\mathrm{dec}}_{\mathrm{local}} = 1/q^{\mathrm{enc}}_{\mathrm{local}}$, and the rate-specific scale
$q^{\mathrm{enc}}$ increases monotonically with the rate point $i$ by parameterizing it as an accumulation of positive
increments:

\begin{equation}
q^{\mathrm{enc}}_{i} = 0.025 + (40 - 0.025)\,\mathrm{sigmoid}\Bigl(q^{\mathrm{raw}}_{0}
  + \sum_{j=1}^{i} \mathrm{softplus}\bigl(q^{\mathrm{raw}}_{j}\bigr)\Bigr),
\end{equation}

with learnable parameters $q^{\mathrm{raw}}_{j}$ and the rescaled sigmoid keeps $q^{\mathrm{enc}}_{i}$ within $(0.025, 40)$
to span a wide range of quantization levels.
With this modification, the encoder and decoder are rate-agnostic, the information loss is concentrated at a single
rounding step, and the element-wise bound of \eqref{eq:latent_bound} holds.
We also verify this empirically: without rounding, $f^{\mathrm{dec}}_\theta(f^{\mathrm{enc}}_\phi(x))$ reaches 40.4\,dB /
41.1\,dB PSNR on Kodak / CLIC2020-test at every evaluated rate point, whereas the unrounded decode of \cite{jia2025towards} drops
to 26.9\,dB / 30.1\,dB at its lowest rate point.
This indicates that its encoder and decoder are rate-dependent and form an information bottleneck at low rates, which
would limit what a compensator in the latent space can recover.
We utilize $q = q^{\mathrm{dec}} \cdot q^{\mathrm{dec}}_{\mathrm{local}}$ when performing the computation with the PerQ
compensator.

\subsection{Training}
\label{sec:method:training}
\textbf{Stage 1: Backbone training.}
We first train the variable-rate compression backbone with
the rate-distortion objective $R + \lambda D$,
following the typical image compression training
pipeline~\cite{balle2018variational}. We then freeze the
encoder, decoder, and entropy models, preserving the
quantization structure of~\eqref{eq:latent_bound} exactly.

\textbf{Stage 2: Compensator training.}
We train the compensator with a combined objective
comprising the rectified-flow loss~\eqref{eq:fm_loss},
distortion and perceptual terms, and an adversarial term:
\begin{equation}
\mathcal{L} = \lambda \bigl(
  \mathcal{L}_{\text{MSE}} + \mathcal{L}_{\text{E-LPIPS}}
  \bigr) + \lambda_{\text{FM}}\mathcal{L}_{\text{FM}} + 
  \lambda_{\text{GAN}} \mathcal{L}_{\text{GAN}},
\end{equation}
where $\mathcal{L}_{\text{E-LPIPS}}$ is the E-LPIPS perceptual
distance~\cite{zhang2018unreasonable, kettunen2019lpips}, $\mathcal{L}_{FM}$ is the flow matching loss \cite{liu2022flow}, and
$\mathcal{L}_{\text{GAN}}$ is a generator loss against a
patch discriminator conditioned on the rate point
(cf.~\cite{mentzer2020high, yi2026rethinking}). Since
the backbone is frozen, no rate term appears in the
Stage 2 objective.

\section{Experimental Configurations}

\noindent\textbf{Implementation Details.}
 
We use the intra-frame model of DCVC-RT~\cite{jia2025towards} as our default compression backbone, supporting 64 rate points, with the architectural modifications described in Section~\ref{sec:method:backbone}. We trained it from scratch on the MLIC training dataset ~\cite{jiang2023mlic, jiang2025mlic++}, with a batch size of 64 for 1M steps in the first stage (backbone training) and a batch size of 16 for 200K steps in the second stage (compensator training). More training details and configurations are provided in the \autoref{app:Implementation}.

\noindent\textbf{Evaluation.}
We evaluate PerQ on Kodak \cite{kodak1993} (24 images at $768\times512$),
CLIC2020-test \cite{toderici2020clic} (428 images at $\sim$2K resolution). We measure pixel fidelity using PSNR in RGB space, and
perceptual quality using LPIPS \cite{zhang2018unreasonable},
DISTS \cite{ding2020image}, and FID \cite{heusel2017gans} computed on $256{\times}256$ patches following \cite{muckley2023improving}.

\noindent\textbf{Benchmark Methods.}
We compare against four baseline categories. The conventional
reference is VVC (VTM-22.0
intra) \cite{bross2021overview}. For distortion-optimal
learned codecs we include ELIC \cite{he2022elic} (PSNR
anchor), EVC \cite{wang2023evc} and DCVC-RT intra~\cite{jia2025towards}. For GAN-based generative
codecs we include MS-ILLM \cite{muckley2023improving}
(perceptual anchor), HiFiC \cite{mentzer2020high}, and
Control-GIC \cite{li2024once}. For diffusion-based
generative codecs, we include PerCo \cite{careil2023towards},
CDC \cite{yang2023lossy}, DiffEIC \cite{li2024toward},
StableCodec \cite{zhang2025stablecodec}, and
OSCAR \cite{guo2025oscar}. All neural baselines are re-run from publicly available checkpoints. Reconstructions are generated on a single
NVIDIA-4090 GPU, except for DiffEIC \cite{li2024toward},
whose 50-step decoding is prohibitively slow on the 4090
and is therefore generated on NVIDIA-H200. Decoding latency
and complexity reported in \autoref{tab:positioning} are
measured on a single NVIDIA-4090 GPU for fair comparison.

\section{Results and Discussion}
\label{sec:experiments:main}
\begin{figure}[!t]
    \centering
    \includegraphics[width=1\linewidth]{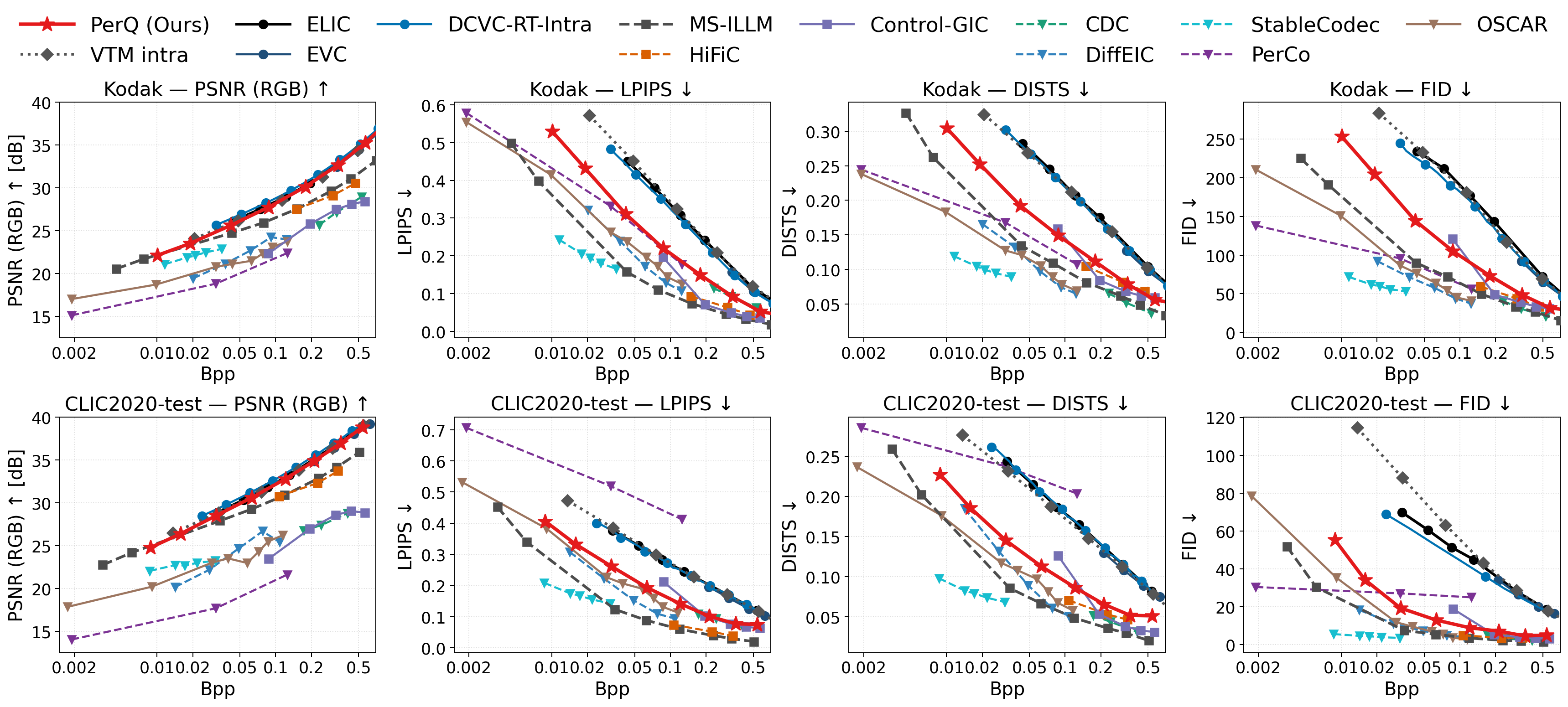}
    \caption{\textbf{Rate--distortion curves on Kodak (top) and
    CLIC2020-test (bottom).} Panels report PSNR (RGB), LPIPS (AlexNet),
    DISTS, and FID across bitrates. PerQ
    (\textcolor{red}{red star}) traces a single curve from one
    trained checkpoint by varying the rate parameter at the decoder.
    Solid markers denote multi-rate or distortion-optimal baselines;
    dashed markers denote per-rate trained generative codecs.}
    \label{fig:rd_curves}
    \vspace{-20pt}
\end{figure}

\noindent\textbf{Quantitative comparison.}
\autoref{tab:positioning} and \autoref{fig:rd_curves} position PerQ across the deployment Pareto frontier on the Kodak and CLIC2020-test datasets. With a single checkpoint, PerQ performs close to ELIC on PSNR ($-1.6\%$ vs.\ $-2.0\%$ BD-rate against VVC on Kodak, $+2.7\%$ vs.\ $-1.9\%$ on CLIC2020-test) while achieving large perceptual gains ($-57.4\%$ in LPIPS, $-63.2\%$ in DISTS and $-67.9\%$ in FID on Kodak, and $-70.6\%$, $-77.4\%$ and $-91.9\%$ on CLIC2020-test). As shown in \autoref{fig:teaser} (right), PerQ is the only codec that combines such perceptual gains with VVC-level PSNR, whereas existing perceptual codecs offer significantly inferior PSNR performance ($+89.9\%$ BD-rate against VVC for MS-ILLM and $+116.3\%$ for HiFiC on CLIC2020-test). It can also be observed that per-rate trained generative codecs achieve higher perceptual BD-rate gains (e.g., MS-ILLM, StableCodec), but they require separate checkpoints per rate point or operate over narrower rate ranges.

\begin{figure}[!t]
    \centering
    \includegraphics[width=0.95\linewidth]{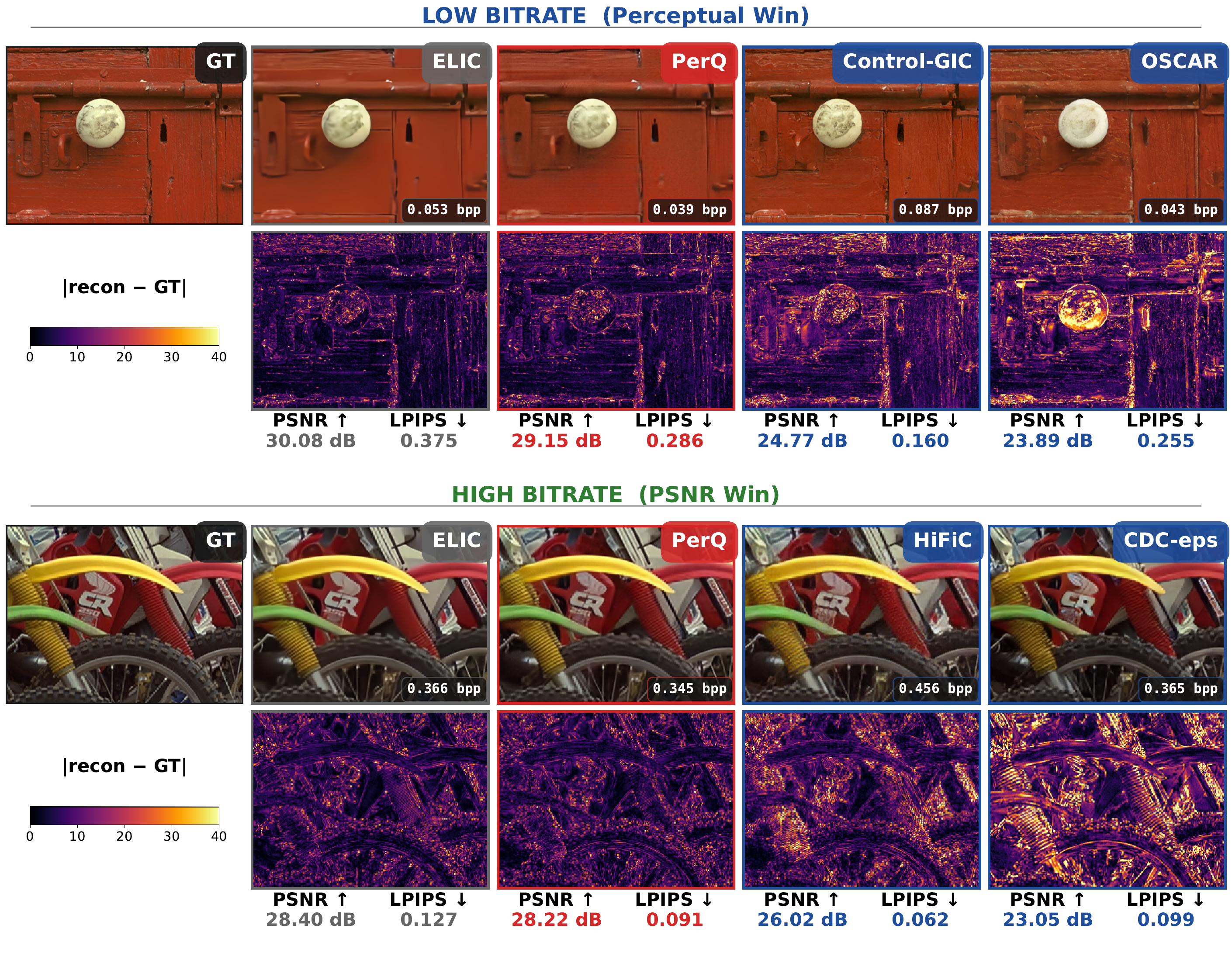}
    \caption{\textbf{Qualitative comparison.}
    Source crop (left) compared with reconstructions from a
    distortion-optimized VAE-based codec (\textcolor{gray}{grey}), PerQ
    (\textcolor{red}{red}, ours), and two generative codecs, GAN-based and diffusion-based
    (\textcolor{blue}{blue}), at similar bitrates with the error maps.
    }

    \label{fig:visual}
    \vspace{-15pt}
\end{figure}

\noindent\textbf{Qualitative comparison.}
\autoref{fig:teaser} (left) and \autoref{fig:visual} compare reconstructions at comparable bitrates. At low rates, ELIC produces visibly blurred textures, while generative codecs recover sharpness but with a substantial PSNR drop (more than $5$~dB below ELIC). PerQ recovers much of this sharpness with PSNR results close to ELIC. In \autoref{fig:teaser}, for example, it matches the LPIPS of OSCAR with a $6.5$~dB higher PSNR. At high rates, generative codecs introduce unfaithful texture details, while PerQ delivers sharpness with similarly high fidelity to ELIC ($28.22$ vs.\ $28.40$~dB) using the fewest bits. This pattern confirms the quantitative results in \autoref{tab:positioning} and verifies the effectiveness of the bounded compensator. More visual examples are available in \autoref{app:visual}.

\noindent\textbf{Ablation Study.}
We ablate the design choices of PerQ on the CLIC2020-test dataset~\cite{toderici2020clic}. We study \textbf{\textit{the compensator design}}, \textbf{\textit{the different backbone}}, and \textbf{\textit{the bound size}} that sets the width of the sub-cell the compensator can reach. Specifically, \autoref{tab:ablation_compensator} ablates four components of the default PerQ design. All variants share the frozen backbone and therefore decode identical bitstreams. It can be observed that replacing E-LPIPS with LPIPS (\textbf{v1.1}) slightly decreases performance, and reducing the E-LPIPS weight to $0.25$ (\textbf{v1.3}) increases the FID BD-rate by $7.20\%$, while a weight of $0.5$ (\textbf{v1.2}) performs similarly to the default. Reducing flow-matching from four integration steps to one (\textbf{v2.1}) degrades perception ($+7.96\%$ LPIPS, $+5.63\%$ DISTS, $+14.75\%$ FID), and lowering $\lambda_{\text{FM}}$ (\textbf{v3.1}, \textbf{v3.2} and \textbf{v3.3}) degrades all metrics, by up to $+28.31\%$ in FID. Finally, the $0.5\times$ compensator ($18.1$M parameters, \textbf{v4.1}) performs close to the default (at most $+3.38\%$, in FID), while the $0.25\times$ one ($8.8$M, \textbf{v4.2}) loses perceptual quality ($+8.85\%$ LPIPS, $+15.64\%$ FID).

The bound scale $m$ in \eqref{eq:clamp} sets the width of the sub-cell and thus the trade-off between perceptual quality and pixel fidelity (\autoref{tab:ablation_bound}). Larger $m$ increases the PSNR cost, while the perceptual savings saturate around $m = 0.2$. Removing the bound brings no further gain. This reflects the geometric role of the transport bound, and the default $m = 0.1$ balances perceptual gains with fidelity while still reaching a peak PSNR of $38.81$~dB.

\autoref{tab:ablation_backbone} further attaches the same compensator to two other frozen backbones, ELIC~\cite{he2022elic} and MLIC++~\cite{jiang2025mlic++}, both retrained for variable-rate coding. The compensator brings similar gains on all three backbones, reducing the LPIPS BD-rate by $42$--$45$ points on Kodak and by $61$--$64$ points on CLIC2020-test at a similar PSNR cost. These consistent gains suggest that PerQ generalizes across backbone architectures.


\begin{table}[!t]
\centering
\caption{\textbf{PerQ design ablation on CLIC2020-test.} BD-rate against the default setting (DCVC-RT intra backbone, 31.0M parameters (the compensator), $4$ steps, $\lambda_{\text{FM}} = 10^{4}$ and trained with the patch-GAN and E-LPIPS perceptual loss (weight $1.0$)).}
\label{tab:ablation_compensator}
\footnotesize
\setlength{\tabcolsep}{4pt}
\begin{tabular}{l  c c c c c}
\toprule
 Configuration & PSNR $\uparrow$ & MS-SSIM $\uparrow$ & LPIPS $\downarrow$ & DISTS $\downarrow$ & FID $\downarrow$ \\
\midrule
\rowcolor{Group2Color!20}
\multicolumn{6}{l}{\textit{Loss setting}} \\
v1.1 LPIPS instead of E-LPIPS & +1.92\% & +1.16\% & +2.85\% & +1.39\% & +5.13\% \\
v1.2 E-LPIPS weight $0.5$     & +0.24\% & +0.97\% & +1.38\% & -2.57\% & -1.60\% \\
v1.3 E-LPIPS weight $0.25$    & +0.61\% & +2.02\% & +3.49\% & +1.15\% & +7.20\% \\
\midrule
\rowcolor{Group3Color!20}
\multicolumn{6}{l}{\textit{Flow matching steps}} \\
v2.1 $1$ step                 & -1.06\% & -1.83\% & +7.96\% & +5.63\% & +14.75\% \\
\midrule
\rowcolor{Group4Color!20}
\multicolumn{6}{l}{\textit{Flow matching loss weight}} \\
v3.1 $\lambda_{\text{FM}} = 10$      & +6.88\% & +5.58\% & +17.28\% & +12.07\% & +27.51\% \\
v3.2 $\lambda_{\text{FM}} = 10^{2}$  & +6.51\% & +5.18\% & +16.29\% & +12.23\% & +28.31\% \\
v3.3 $\lambda_{\text{FM}} = 10^{3}$  & +4.14\% & +2.51\% & +10.59\% & +7.49\% & +22.69\% \\
\midrule
\rowcolor{Group5Color!20}
\multicolumn{6}{l}{\textit{Compensator size}} \\
v4.1 PerQ-$0.5\times$ (18.1M)           & -0.03\% & -0.22\% & -0.59\% & -0.05\% & +3.38\% \\
v4.2 PerQ-$0.25\times$ (8.8M)           & -0.27\% & -1.10\% & +8.85\% & +11.05\% & +15.64\% \\

\bottomrule
\end{tabular}
\vspace{-15pt}
\end{table}

\begin{table}[!t]
\centering
\caption{\textbf{Bound scale ablation on CLIC2020-test.} BD-rate against the unmodified backbone for the bound scale $m$ in \eqref{eq:clamp}. PSNR$_\text{min}$/PSNR$_\text{max}$ report the average PSNR at the lowest/highest rate point.}
\label{tab:ablation_bound}
\footnotesize
\setlength{\tabcolsep}{4pt}
\begin{tabular}{l c c c c c c c}
\toprule
Bound scale $m$ & PSNR $\uparrow$ & MS-SSIM $\uparrow$ & LPIPS $\downarrow$ & DISTS $\downarrow$ & FID $\downarrow$ & PSNR$_\text{min}$ & PSNR$_\text{max}$ \\
\midrule
$0.05$                            & +13.77\% & +16.20\% & -60.56\% & -76.61\% & -91.87\% & 24.88 & 39.16 \\
$0.10$ \textbf{(default)}           & +18.24\% & +20.60\% & -67.84\% & -79.68\% & -91.51\% & 24.76 & 38.81 \\
$0.20$                             & +23.23\% & +24.40\% & -70.79\% & -82.53\% & -92.39\% & 24.41 & 38.47 \\
unbounded                          & +28.58\% & +25.38\% & -64.45\% & -81.73\% & -92.22\% & 24.47 & 38.11 \\
\bottomrule
\end{tabular}

\vspace{-12pt}
\end{table}

\begin{table}[t]
\caption{\textbf{PerQ compensator on three frozen backbones.} The same compensator, bound and training recipe are attached to three
variable-rate backbones (retrained), each covering 64 rate points with one checkpoint. BD-rate (\%) against VVC intra.}
\label{tab:ablation_backbone}
\centering
\resizebox{\linewidth}{!}{\begin{tabular}{l ccccc ccccc}
\toprule
 & \multicolumn{5}{c}{\textbf{Kodak}} & \multicolumn{5}{c}{\textbf{CLIC2020-test}} \\
\cmidrule(lr){2-6} \cmidrule(lr){7-11}
 & bpp & PSNR & LPIPS & DISTS & FID & bpp & PSNR & LPIPS & DISTS & FID \\
\midrule
\rowcolor{Group6Color!20}
\multicolumn{11}{l}{\textit{DCVC-RT backbone~\cite{jia2025towards} (default)}} \\
    Backbone only & 0.010--0.87 & $-10.6$ & $-12.7$ & $-5.7$ & $-10.4$ & 0.009--0.54 & $-11.1$ & $-7.3$ & $+3.5$ & $-42.2$ \\
    \textbf{+ PerQ compensator} & 0.010--0.87 & {\boldmath$-1.6$} & {\boldmath$-57.4$} & {\boldmath$-63.2$} & {\boldmath$-67.9$} & 0.009--0.54 & {\boldmath$+2.7$} & {\boldmath$-70.6$} & {\boldmath$-77.4$} & {\boldmath$-91.9$} \\
\midrule
\rowcolor{Group7Color!20}
\multicolumn{11}{l}{\textit{ELIC backbone~\cite{he2022elic}}} \\
    Backbone only & 0.061--0.91 & $-4.9$ & $-13.8$ & $-5.2$ & $-5.5$ & 0.045--0.56 & $-5.5$ & $-8.5$ & $+4.9$ & $-28.4$ \\
    \textbf{+ PerQ compensator} & 0.061--0.91 & {\boldmath$+4.0$} & {\boldmath$-57.4$} & {\boldmath$-64.4$} & {\boldmath$-64.2$} & 0.045--0.56 & {\boldmath$+9.0$} & {\boldmath$-72.5$} & {\boldmath$-77.8$} & {\boldmath$-88.5$} \\
\midrule
\rowcolor{Group8Color!20}
\multicolumn{11}{l}{\textit{MLIC++ backbone~\cite{jiang2025mlic++}}} \\
    Backbone only & 0.055--0.85 & $-9.1$ & $-20.5$ & $-10.5$ & $-13.0$ & 0.041--0.52 & $-11.0$ & $-15.8$ & $-4.0$ & $-35.4$ \\
    \textbf{+ PerQ compensator} & 0.055--0.85 & {\boldmath$+0.0$} & {\boldmath$-63.0$} & {\boldmath$-68.4$} & {\boldmath$-68.0$} & 0.041--0.52 & {\boldmath$+1.2$} & {\boldmath$-76.9$} & {\boldmath$-79.2$} & {\boldmath$-89.4$} \\
\bottomrule
\end{tabular}}
\vspace{-10pt}
\end{table}


We note that these results are obtained with a frozen backbone and the compensator as an add-on module and more aggressive modification of the compression model is expected to widen the distortion-perception trade-off further. We discuss the limitation in the \autoref{app:limitation}.


\section{Conclusion}
\label{sec:conclusion}

We presented PerQ, a learning-based image codec that addresses the trade-off between pixel-wise fidelity and perceptual quality by formulating perceptual reconstruction as inverse generative modeling with bounded support. Specifically, we characterized the quantization cell of a multi-rate distortion-optimized codec as the set of latents that yield the same transmitted symbols, and derived a transport bound that constrains the deviation of reconstructions within this cell from the backbone's output. Building on this formulation, we developed a rate-conditioned flow-matching compensator that performs perceptual reconstruction within the codec-induced bounded support, preserving the backbone's multi-rate capability. Experiments on Kodak and CLIC2020-test demonstrate that PerQ achieves strong perceptual quality with competitive pixel fidelity, while supporting a wide bitrate range through a single model and efficient decoding. To the best of our knowledge, PerQ is the first generative image codec to combine these properties in a single framework.

\section*{Acknowledgements}
This work was supported by UK EPSRC, the UKRI MyWorld Strength in Places Programme, Tencent (US) and the University of Bristol. Part of the computational work was also supported by the facilities provided by the Advanced Computing Research Centre at the University of Bristol. The authors acknowledge the use of resources provided by the Isambard-AI National AI Research Resource (AIRR) \cite{mcintoshsmith2024isambard}. Isambard-AI is operated by the University of Bristol and is funded by the UK Government’s Department for Science, Innovation and Technology (DSIT) via UK Research and Innovation; and the Science and Technology Facilities Council [ST/AIRR/I-A-I/1023].


\bibliographystyle{abbrv}
\bibliography{neurips_2026}


\newpage

\appendix

{\Large \textbf{Appendix}}

\section{From the quantization cell to the compensated distribution}
\label{app:bounded_support}

This section formalizes the connection between the quantization cell and the distribution induced by the compensator. Throughout, we write $g := f^{\mathrm{dec}}_\theta$ for simplicity.

Let $X$ denote the random variable of source images with distribution $p_X$, let $Y = f^{\mathrm{enc}}_\phi(X)$ be its latent, and let $(\hat Y, \hat Z, Q)$ be the entropy-decoded latent, the hyper-prior latent, and the rate state obtained by quantizing $Y$. The backbone reconstruction is $\hat X = g(\hat Y)$ and the unquantized reconstruction is $X^* := g(Y)$. Although $\hat Y$, $\hat Z$, and $Q$ are deterministic functions of $X$ given the frozen backbone, we treat them as random variables with the joint distribution induced by $p_X$. The joint distribution of the source latent and the decoder inputs factors as
\begin{equation}
  p_{Y, \hat Y, \hat Z, Q}(y, \hat y, \hat z, q)
  =
  p_{Y \mid \hat Y, \hat Z, Q}(y \mid \hat y, \hat z, q)
  p_{\hat Y, \hat Z, Q}(\hat y, \hat z, q),
  \label{eq:joint_factor}
\end{equation}
where $p_{\hat Y, \hat Z, Q}$ is fixed by the frozen backbone and $p_{Y \mid \hat Y, \hat Z, Q}$ is the posterior distribution of the source latent given the decoder inputs. This conditional distribution characterizes the source latents compatible with the decoder inputs, whose support we establish next.

\noindent\textbf{The posterior has support inside the quantization cell.}
The quantization bound \eqref{eq:latent_bound} states that $|Y - \hat Y| \le Q/2$ holds element-wise for every encoded image. Thus, $Y \in \mathcal{C}(\hat Y, Q)$ holds deterministically under $p_{Y, \hat Y, \hat Z, Q}$. Conditioning on $(\hat Y, \hat Z, Q) = (\hat y, \hat z, q)$ gives
\begin{equation}
  \mathrm{supp}\bigl(
  p_{Y \mid \hat Y, \hat Z, Q}(\cdot \mid
  \hat y, \hat z, q)
  \bigr)
  \subseteq \mathcal{C}(\hat y, q).
  \label{eq:cond_support}
\end{equation}
Conversely, every $y' \in \mathcal{C}(\hat y, q)$ quantizes to the same transmitted symbols, up to ties on the cell boundary, given the hyper-prior latent $\hat z$ that determines the step size $q$. The cell therefore characterizes the latents consistent with the transmitted symbols. This makes it a natural support constraint for the compensator: it contains the source latent while restricting generation to symbol-consistent alternatives.

\noindent\textbf{The compensator induces a marginal.}
The compensator defines a conditional distribution $p_\psi(\tilde y \mid \hat y, \hat z, q)$ supported on $\mathcal{C}(\hat y, q)$ for every $(\hat y, \hat z, q)$, as specified in \eqref{eq:cell_constraint}. Combined with the fixed marginal $p_{\hat Y, \hat Z, Q}$, this conditional distribution induces a marginal distribution on the compensated latent $\tilde Y$:
\begin{equation}
  p_{\tilde Y}(\tilde y)
  =
  \int p_\psi(\tilde y \mid \hat y, \hat z, q)
  p_{\hat Y, \hat Z, Q}(\hat y, \hat z, q)
  \, d\hat y \, d\hat z \, dq.
  \label{eq:marg_compensated}
\end{equation}
Through the fixed decoder, it further induces a marginal $p_{\tilde X}$ on the compensated reconstruction $\tilde X = g(\tilde Y)$. Closing the perceptual gap of \cite{blau2019rethinking} ideally requires $p_{\tilde X} = p_X$. We next characterize the distribution obtained by matching the source-latent posterior, providing a reference for the compensator's achievable perceptual improvement under the frozen backbone.

\noindent\textbf{Conditional matching reproduces the unquantized reconstruction.}
Since $p_{\hat Y, \hat Z, Q}$ is fixed, if
\begin{equation}
  p_\psi(\tilde y \mid \hat y, \hat z, q)
  =
  p_{Y \mid \hat Y, \hat Z, Q}(\tilde y \mid \hat y, \hat z, q)
  \quad
  \text{for } p_{\hat Y, \hat Z, Q}\text{-almost every }
  (\hat y, \hat z, q),
  \label{eq:cond_matching}
\end{equation}
then substituting \eqref{eq:cond_matching} into \eqref{eq:marg_compensated} yields $p_{\tilde Y} = p_Y$ by \eqref{eq:joint_factor}. Since $g$ is fixed, it follows that $p_{\tilde X} = p_{X^*}$. Thus, matching the latent posterior reproduces the distribution of the unquantized reconstruction $X^* = g(Y)$, rather than necessarily recovering the source-image distribution $p_X$. The remaining divergence $d_p(p_X, p_{X^*})$ arises from the backbone transform itself and remains fixed once the backbone is frozen. The compensator instead targets the rate-dependent component of the perceptual gap, which generally increases as the rate decreases. Several existing generative codecs also pursue conditional distribution matching, typically conditioning on quantized latents \cite{yang2023lossy, careil2023towards} or learned semantic codes \cite{li2024toward, careil2023towards}. PerQ conditions on $(\hat y, \hat z, q)$ and constrains the conditional distribution to the quantization cell, yielding the bounded support in \eqref{eq:cond_support}.

\noindent\textbf{The decoded samples stay inside $\mathcal{B}(\hat x, q)$.}
Beyond this distributional result, the support constraint also provides a geometric guarantee on the compensated reconstructions through the transport bound. For every $\tilde y \in \mathcal{C}(\hat y, q)$, \eqref{eq:transport} gives
\[
  \|g(\tilde y) - g(\hat y)\|_2 \le R(q).
\]
Hence, every sample from $p_\psi$ decodes into $\mathcal{B}(\hat x, q)$, where $\hat x = g(\hat y)$. With the clamp in \eqref{eq:clamp}, the decoded sample lies within the smaller ball of radius $mR(q)$. In particular, $X^* = g(Y) \in \mathcal{B}(\hat X, Q)$. This bound limits how far the compensator can move a reconstruction from the backbone output, allowing perceptual refinement within a rate-dependent bounded region.

\section{Implementation details}
\label{app:Implementation}
\subsection{Training configurations}
\paragraph{Stage 1: backbone (distortion only).}
The compression backbone (DCVC-RT intra-frame model \cite{jia2025towards}) is trained from scratch on the MLIC training
set~\cite{jiang2023mlic, jiang2025mlic++} with the rate-distortion objective $L = R + \lambda \cdot L_{\text{MSE}}$ for
1M steps at batch size 64.
The Lagrange multiplier $\lambda$ is sampled from [4, 4096] for 64 rate points, following \cite{li2024neural}.

We use the AdamW optimizer~\cite{loshchilov2017decoupled} with $(\beta_1, \beta_2) = (0.9, 0.999)$ and weight decay
$10^{-2}$.
The peak learning rate is $2 \times 10^{-4}$, reached by a linear warmup from $10^{-6}$ over the first 100K steps, with
a multi-step schedule that halves the learning rate at $50\%$, $75\%$, $87.5\%$, and $93.75\%$ of training.
Gradients are clipped to a global $\ell_2$ norm of 1.0.
Over the first $10\%$ of training, the weight of the rate term and $\lambda$ are ramped up linearly, from 0 to 1 and
from 4 to the sampled value, respectively. 
Training images are randomly cropped to $256{\times}256$.

\paragraph{Stage 2: compensator (perceptual optimization).}
In stage 2 we freeze the encoder, decoder, hyper-prior, and entropy models, and add the compensator and discriminator
for joint training.
We train for 200K steps at batch size 16 with $\lambda \in [4, 4096]$ and the Stage 2 objective of
Section~\ref{sec:method:training}.
The weight of the latent flow-matching term is $\lambda_{\text{FM}} = 10^{4}$, the weight of the adversarial term is
$\lambda_{\text{GAN}} = 0.1$, and the E-LPIPS term has the same weight as the MSE term.

$L_{\text{E-LPIPS}}$ uses the VGG backbone~\cite{kettunen2019lpips, zhang2018unreasonable, simonyan2014very} during
training; we report the LPIPS variant with AlexNet \cite{krizhevsky2012imagenet} features for evaluation.
Optimizer and schedule follow stage 1, except the peak learning rate is $1 \times 10^{-4}$, the warmup is 10K steps
without the rate and $\lambda$ warmup, and we keep the weight decay to $10^{-2}$ on the compensator's and discriminator's
parameters.
The GAN term is disabled for the first 20K iterations and then linearly warmed up over the next 20K iterations to its full
weight.

\paragraph{Compensator architecture.}
The velocity field $v_\psi$ is a stack of 10 ConvNeXt blocks~\cite{liu2022convnet} with hidden width of 512, an MLP
expansion ratio of 4 and a kernel size of 5.
It first computes the sum of a $1{\times}1$ convolution of the scaled flow state $q\,r_t$ and a $1{\times}1$ convolution of
the channel-wise concatenation of $\hat y$, the upsampled $\hat z$, and $q$.
Timestep conditioning is injected through a sinusoidal positional embedding followed by a linear
layer with SiLU, and each block maps it through a zero-initialized linear layer to the shift, scale, and gate of an
adaLN-zero modulation~\cite{peebles2023scalable}.
The output head is a LayerNorm followed by a zero-initialized $1{\times}1$ convolution, which makes the compensator output only
the bounded prior noise at the start of training, i.e., it acts as the identity up to this noise.
We use 4 Euler steps for flow matching at both training and inference, with bound scale $m{=}0.1$ (Eq.~\ref{eq:clamp})
and prior noise standard deviation $\sigma = 0.029 \approx m/\sqrt{12}$ (Section~\ref{sec:compensator_arch}).

The default compensator has 31.0M parameters.
The compensator-size ablation (\textbf{v4.1} and \textbf{v4.2} in \autoref{tab:ablation_compensator}) uses two smaller
compensators that change only the hidden width and keep the depth, kernel size, bound and training recipe:
PerQ-$0.5\times$ with a hidden width of 384 (18.1M parameters) and PerQ-$0.25\times$ with a hidden width of 256 (8.8M
parameters).

\paragraph{Discriminator.}
The adversarial term uses a multi-scale rate-point-conditioned patch discriminator \cite{mentzer2020high} with 3 spatial
scales (input downsampled by $1{\times}, 2{\times}, 4{\times}$ via average pooling) and a separate per-rate-point
parameterization (per-rate kernels at each conv layer).
Each scale uses 3 strided convolutional layers with base width 48, doubled at each layer (capped at 512), followed by a
$4{\times}4$ stride-1 conv and a final $4{\times}4$ patch logit head.
All convolutions except the final logit head use spectral normalization~\cite{miyato2018spectral} and LeakyReLU($0.2$)
activations.
The discriminator is trained jointly with the compensator, where both are updated at every step by the same AdamW optimizer,
with hinge losses for the discriminator and the generator.

\paragraph{Backbones of the generality study.}
The three backbones of \autoref{tab:ablation_backbone} are all trained from scratch on the MLIC training set with the
Stage 1 objective, 64 rate points and the consolidated bottleneck quantization of Section~\ref{sec:method:backbone}.
DCVC-RT is the default backbone described above.
ELIC~\cite{he2022elic} keeps its original architecture (320 latent channels in five uneven groups) and follows the
DCVC-RT recipe, except that $\lambda$ is sampled from $[32, 4096]$.
MLIC++~\cite{jiang2025mlic++} keeps its original architecture (320 latent channels in 10 slices) with $\lambda \in [32,
4096]$.
It is first trained for 1M steps on $256{\times}256$ crops at batch size 64 with a peak learning rate of $10^{-4}$, and
then fine-tuned for 400K steps on $512{\times}512$ crops at batch size 32,
with a peak learning rate of $10^{-4}$, 5K warmup steps, no rate or $\lambda$ warmup, and the learning rate halved at
$37.5\%$, $75\%$, $87.5\%$, and $93.75\%$ of the fine-tuning.
Since $\lambda \geq 32$, ELIC and MLIC++ do not reach the lowest rates of DCVC-RT (see the bpp ranges in
\autoref{tab:ablation_backbone}).
On each backbone, the compensator and all Stage 2 settings are those of the default model, except that $\lambda$ follows
the range of the backbone and MLIC++ is trained on $512{\times}512$ crops.

\section{Additional quantitative results}
\label{app:kid}
\begin{figure}[htbp]
\centering
\includegraphics[width=1\linewidth]{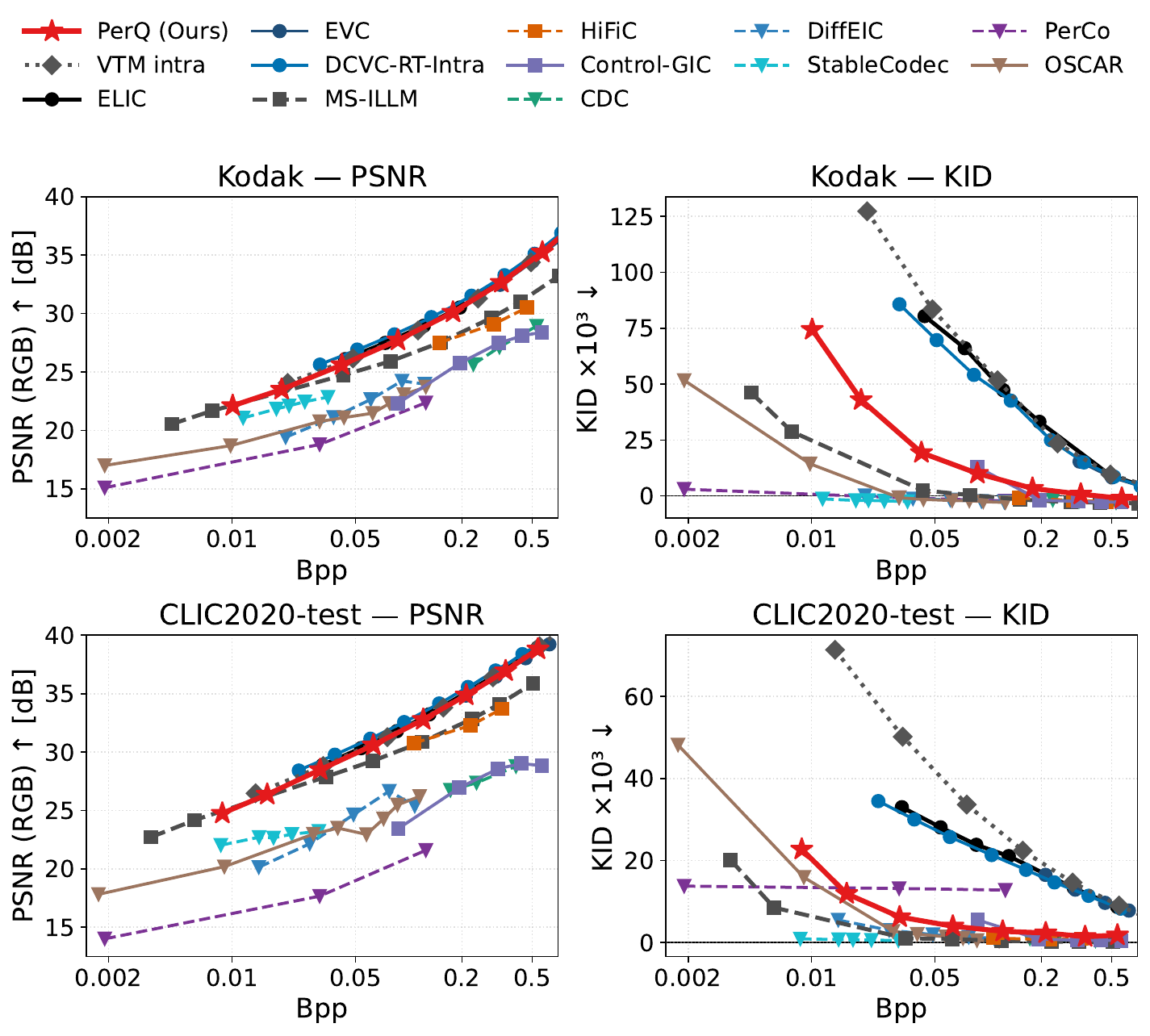}
\caption{\textbf{Rate-PSNR and rate-KID curves on Kodak (top) and CLIC2020-test (bottom).} KID ($\times 10^{3}$, lower is better) is computed on the same $256{\times}256$ patches as FID. }
\label{fig:app_rd_kid}
\end{figure}

Tables~\ref{tab:app_bd_kodak} and \ref{tab:app_bd_clic} extend Table~\ref{tab:positioning} with the bitrate range of every codec and with the BD-rate in KID~\cite{binkowski2018demystifying}, computed on the same $256{\times}256$ patches as FID. Since KID uses an unbiased estimator, it is less affected than FID by the small number of patches on Kodak ($192$), although its estimate can fall slightly below zero when the reconstructions are statistically close to the source. \autoref{fig:app_rd_kid} shows the corresponding rate-KID curves next to the rate-PSNR curves. The KID results follow those of FID: on Kodak, PerQ ($-84.1\%$) is within $2.5$ points of HiFiC ($-86.6\%$) and Control-GIC ($-85.7\%$), and on CLIC2020-test it surpasses Control-GIC ($-95.0\%$ vs.\ $-90.8\%$), while keeping its PSNR BD-rate within $3\%$ of VVC.

\begin{table}[htbp]
\caption{\textbf{BD-rate against VVC on Kodak (\%, lower is better) with bitrate ranges and KID.} Same anchor, fitting method and N/A rule as Table~\ref{tab:positioning}. The bpp range covers the rate points evaluated for each codec.}
\label{tab:app_bd_kodak}
\centering
\small
\setlength{\tabcolsep}{4pt}
\begin{tabular}{l c c r r r r r}
\toprule
Codec & \shortstack[c]{Single-model\\ multi-rate} & bpp range & PSNR & LPIPS & DISTS & FID & KID \\
\midrule
\multicolumn{8}{l}{\textit{Conventional codec}} \\
    VTM intra~\cite{bross2021overview} &  & 0.113--2.316 & 0 & 0 & 0 & 0 & 0 \\
\midrule
\multicolumn{8}{l}{\textit{VAE-based codecs}} \\
    ELIC~\cite{he2022elic} &  & 0.044--0.865 & $-2.0$ & $-2.1$ & $+2.4$ & $+7.3$ & $+3.8$ \\
    EVC~\cite{wang2023evc} & \cmark & 0.329--0.983 & $+2.0$ & $-9.0$ & $-5.4$ & $-4.0$ & $-7.5$ \\
    DCVC-RT-Intra~\cite{jia2025towards} & \cmark & 0.031--0.913 & $-11.2$ & $-8.7$ & $-2.3$ & $-1.8$ & $-2.8$ \\
\midrule
\multicolumn{8}{l}{\textit{GAN-based perceptual codecs}} \\
    MS-ILLM~\cite{muckley2023improving} &  & 0.0045--0.708 & $+85.9$ & $-86.4$ & $-82.9$ & $-86.3$ & $-95.9$ \\
    HiFiC~\cite{mentzer2020high} &  & 0.149--0.468 & $+128.6$ & $-69.8$ & $-59.0$ & $-64.3$ & $-86.6$ \\
    Control-GIC$^{\S}$~\cite{li2024once} & \cmark & 0.087--0.568 & $+372.4$ & $-74.0$ & $-68.5$ & $-69.7$ & $-85.7$ \\
\midrule
\multicolumn{8}{l}{\textit{Diffusion-based perceptual codecs}} \\
    PerCo~\cite{careil2023towards} &  & 0.0019--0.125 & N/A & $-64.1$ & $-84.4$ & $-92.9$ & $-99.0$ \\
    DiffEIC~\cite{li2024toward} &  & 0.020--0.123 & N/A & $-81.0$ & $-88.8$ & $-91.3$ & $-97.6$ \\
    CDC~\cite{yang2023lossy} &  & 0.232--0.530 & $+318.7$ & $-48.6$ & $-71.7$ & $-68.8$ & $-85.9$ \\
    StableCodec~\cite{zhang2025stablecodec} &  & 0.012--0.035 & N/A & $-92.7$ & $-96.2$ & $-96.8$ & $-99.0$ \\
    OSCAR~\cite{guo2025oscar} & \cmark & 0.0019--0.125 & N/A & $-79.0$ & $-91.8$ & $-92.3$ & $-97.8$ \\
\midrule
    \textbf{PerQ (Ours)} & \cmark & 0.010--0.875 & {\boldmath$-1.6$} & {\boldmath$-57.4$} & {\boldmath$-63.2$} & {\boldmath$-67.9$} & {\boldmath$-84.1$} \\
\bottomrule
\end{tabular}
\end{table}

\begin{table}[htbp]
\caption{\textbf{BD-rate against VVC on CLIC2020-test (\%, lower is better) with bitrate ranges and KID.} Same protocol as Table~\ref{tab:app_bd_kodak}.}
\label{tab:app_bd_clic}
\centering
\small
\setlength{\tabcolsep}{4pt}
\begin{tabular}{l c c r r r r r}
\toprule
Codec & \shortstack[c]{Single-model\\ multi-rate} & bpp range & PSNR & LPIPS & DISTS & FID & KID \\
\midrule
\multicolumn{8}{l}{\textit{Conventional codec}} \\
    VTM intra~\cite{bross2021overview} &  & 0.076--1.717 & 0 & 0 & 0 & 0 & 0 \\
\midrule
\multicolumn{8}{l}{\textit{VAE-based codecs}} \\
    ELIC~\cite{he2022elic} &  & 0.032--0.538 & $-1.9$ & $-1.3$ & $+9.4$ & $-13.8$ & $-31.6$ \\
    EVC~\cite{wang2023evc} & \cmark & 0.211--0.626 & $+3.7$ & $-8.7$ & $-2.3$ & $-6.2$ & $-11.9$ \\
    DCVC-RT-Intra~\cite{jia2025towards} & \cmark & 0.024--0.562 & $-11.8$ & $-2.5$ & $+8.8$ & $-30.1$ & $-39.7$ \\
\midrule
\multicolumn{8}{l}{\textit{GAN-based perceptual codecs}} \\
    MS-ILLM~\cite{muckley2023improving} &  & 0.0035--0.506 & $+89.9$ & $-91.0$ & $-90.7$ & $-97.4$ & $-98.7$ \\
    HiFiC~\cite{mentzer2020high} &  & 0.107--0.336 & $+116.3$ & $-81.1$ & $-76.9$ & $-83.0$ & N/A \\
    Control-GIC$^{\S}$~\cite{li2024once} & \cmark & 0.087--0.567 & $+1045.8$ & $-65.6$ & $-73.8$ & $-85.2$ & $-90.8$ \\
\midrule
\multicolumn{8}{l}{\textit{Diffusion-based perceptual codecs}} \\
    PerCo~\cite{careil2023towards} &  & 0.0019--0.125 & N/A & N/A & N/A & $-94.3$ & $-94.5$ \\
    DiffEIC~\cite{li2024toward} &  & 0.014--0.108 & N/A & $-84.6$ & $-87.4$ & $-96.5$ & $-97.6$ \\
    CDC~\cite{yang2023lossy} &  & 0.172--0.405 & N/A & $-63.3$ & $-74.7$ & $-86.8$ & N/A \\
    StableCodec~\cite{zhang2025stablecodec} &  & 0.0086--0.031 & N/A & $-94.9$ & $-97.1$ & $-99.0$ & N/A \\
    OSCAR~\cite{guo2025oscar} & \cmark & 0.0018--0.115 & N/A & $-80.2$ & $-88.7$ & $-96.0$ & $-96.8$ \\
\midrule
    \textbf{PerQ (Ours)} & \cmark & 0.0088--0.539 & {\boldmath$+2.7$} & {\boldmath$-70.6$} & {\boldmath$-77.4$} & {\boldmath$-91.9$} & {\boldmath$-95.0$} \\
\bottomrule
\end{tabular}
\end{table}


\section{More visual results}
\label{app:visual}

\begin{figure}[t]
    \centering
    \includegraphics[width=1\linewidth]{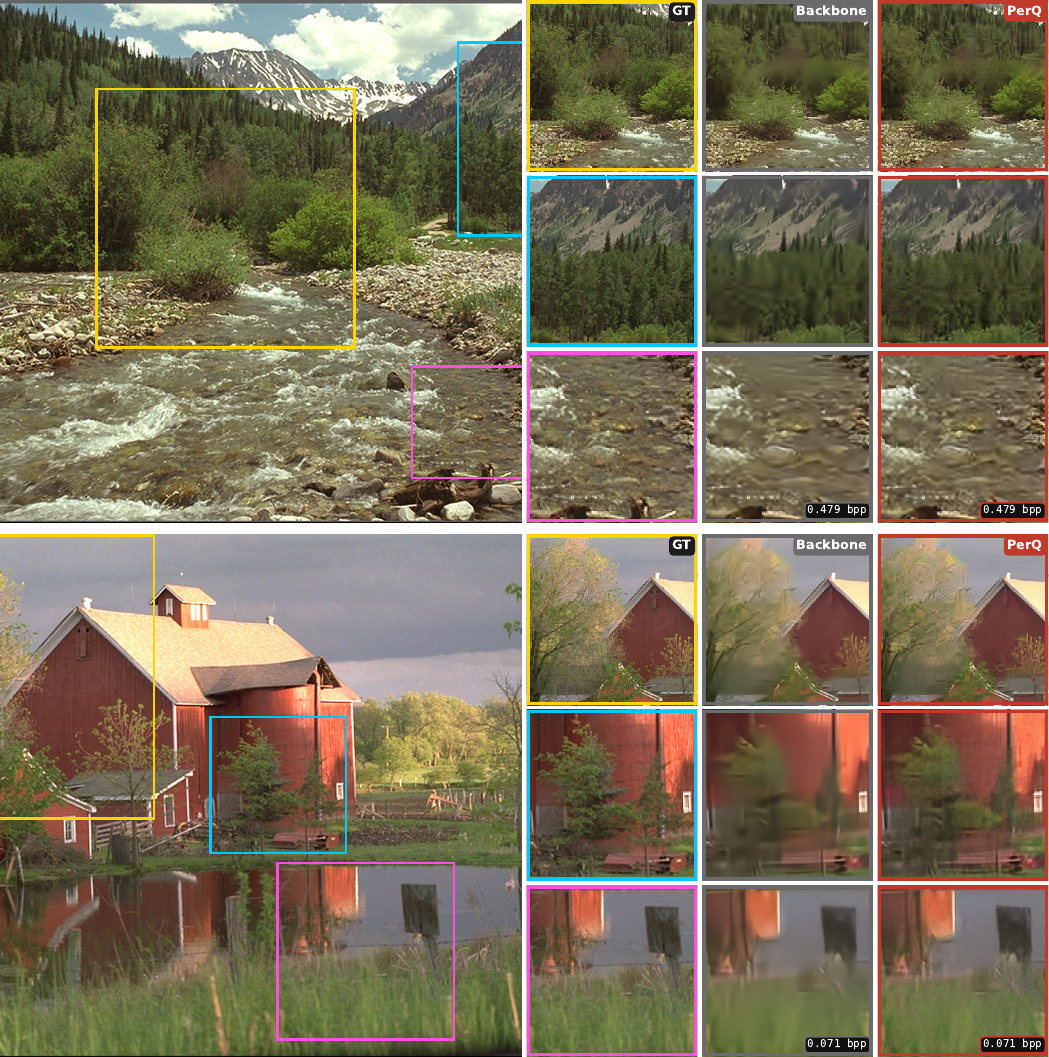}
    \caption{\textbf{Visual comparison between the backbone and PerQ on Kodak}, with the DCVC-RT (top) and MLIC++ (bottom) backbones. Additional examples on CLIC2020-test are shown in \autoref{fig:visual_clic_dcvc} (DCVC-RT) and \autoref{fig:visual_clic_mlic} (MLIC++).}
    \label{fig:visual_kodak}
\end{figure}

\begin{figure}[t]
    \centering
    \includegraphics[width=1\linewidth]{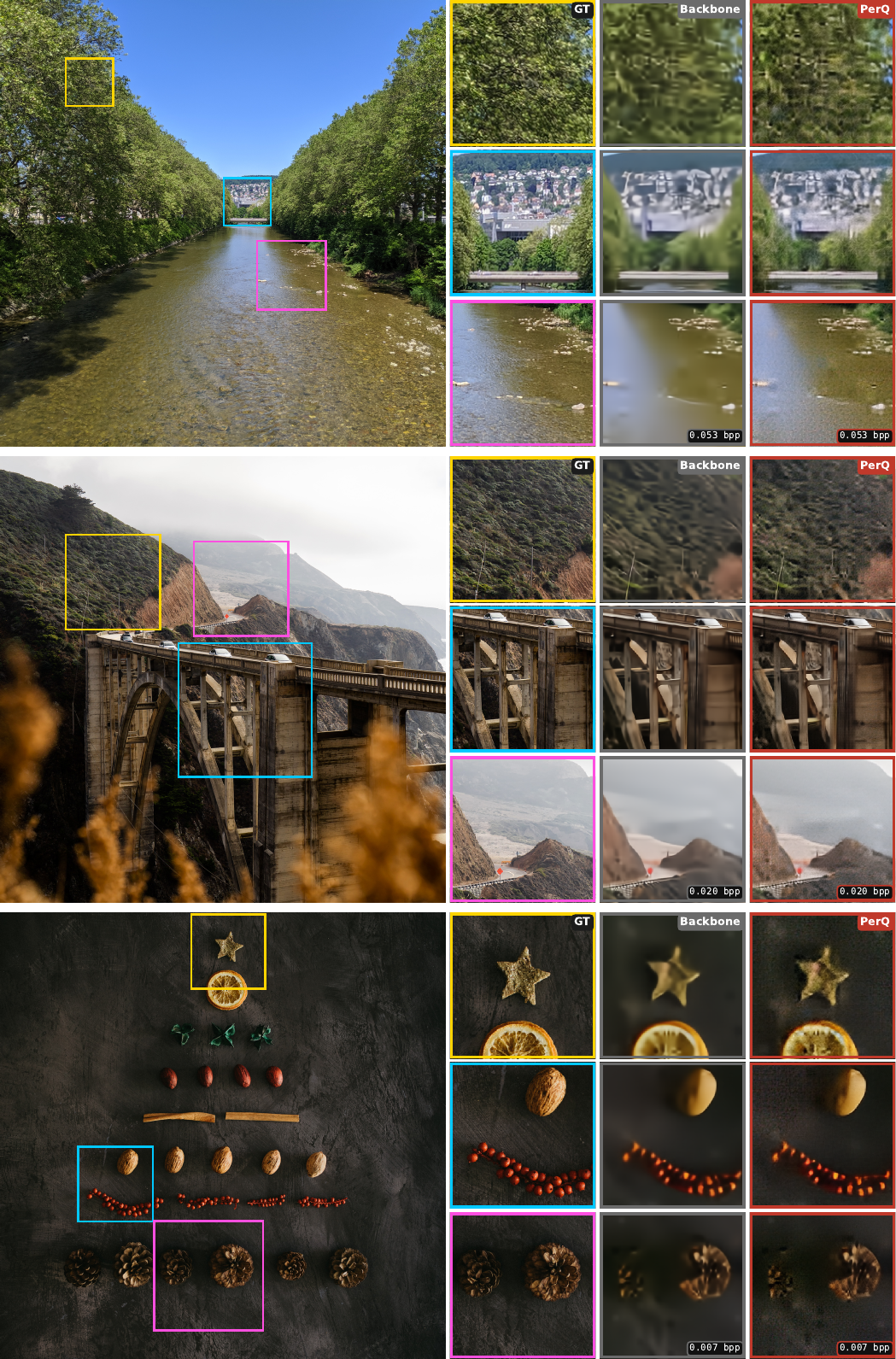}
    \caption{\textbf{Visual comparison between the backbone and PerQ on CLIC2020-test} with the DCVC-RT backbone.}
    \label{fig:visual_clic_dcvc}
\end{figure}

\begin{figure}[t]
    \centering
    \includegraphics[width=1\linewidth]{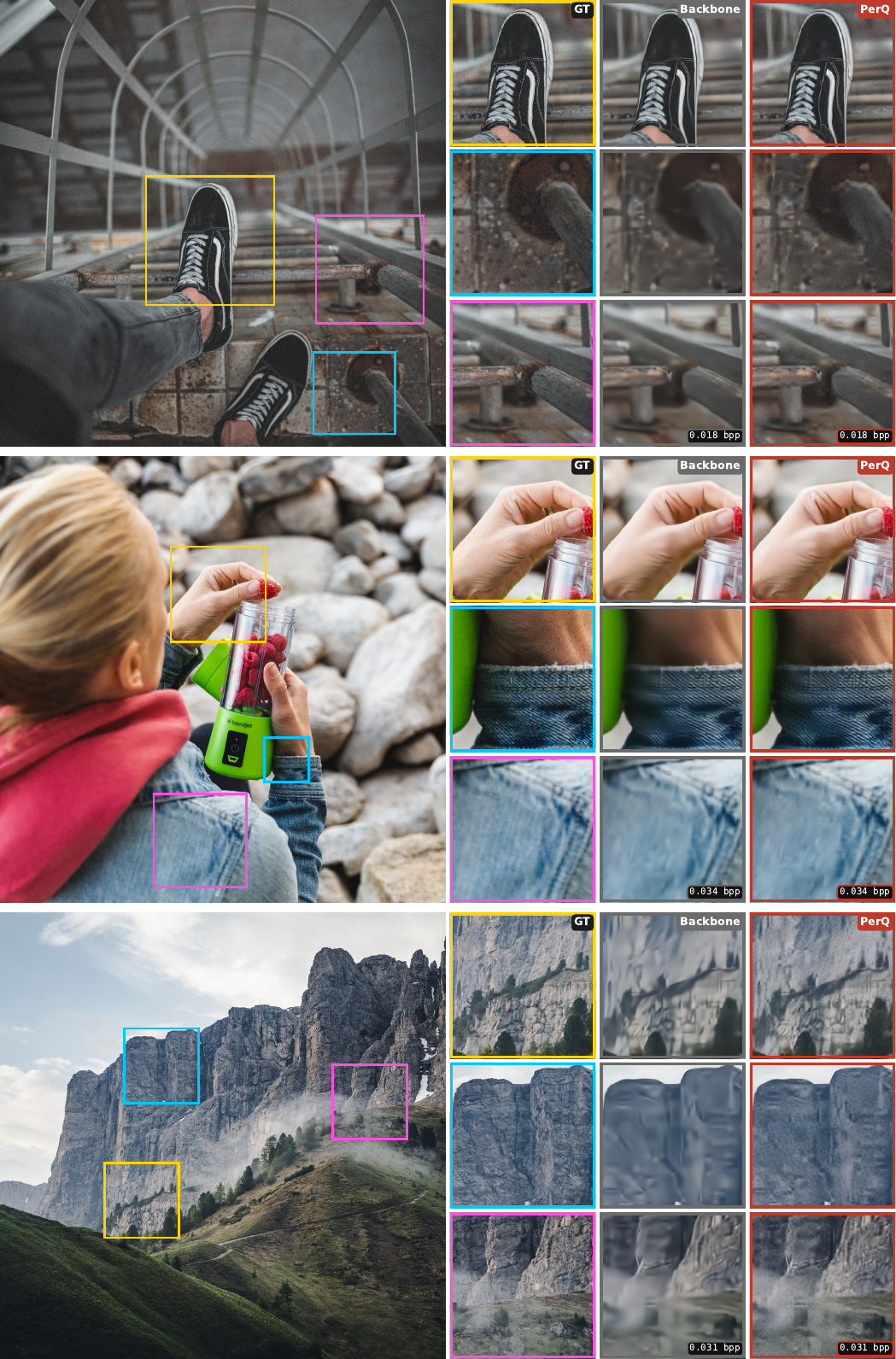}
    \caption{\textbf{Visual comparison between the backbone and PerQ on CLIC2020-test} with the MLIC++ backbone of the generality study (\autoref{tab:ablation_backbone}).}
    \label{fig:visual_clic_mlic}
\end{figure}

Figures~\ref{fig:visual_kodak}, \ref{fig:visual_clic_dcvc}, and~\ref{fig:visual_clic_mlic} compares the visual quality of the backbone models (DCVC-RT \cite{jia2025towards} and MLIC++ \cite{jiang2025mlic++}) with PerQ. In all examples, PerQ significantly improve the visual quality, like synthesizing the texture which are missing in the original backbone outputs, and improving the the sharpness of the images.

\begin{figure}[t]
    \centering
    \includegraphics[width=1\linewidth]{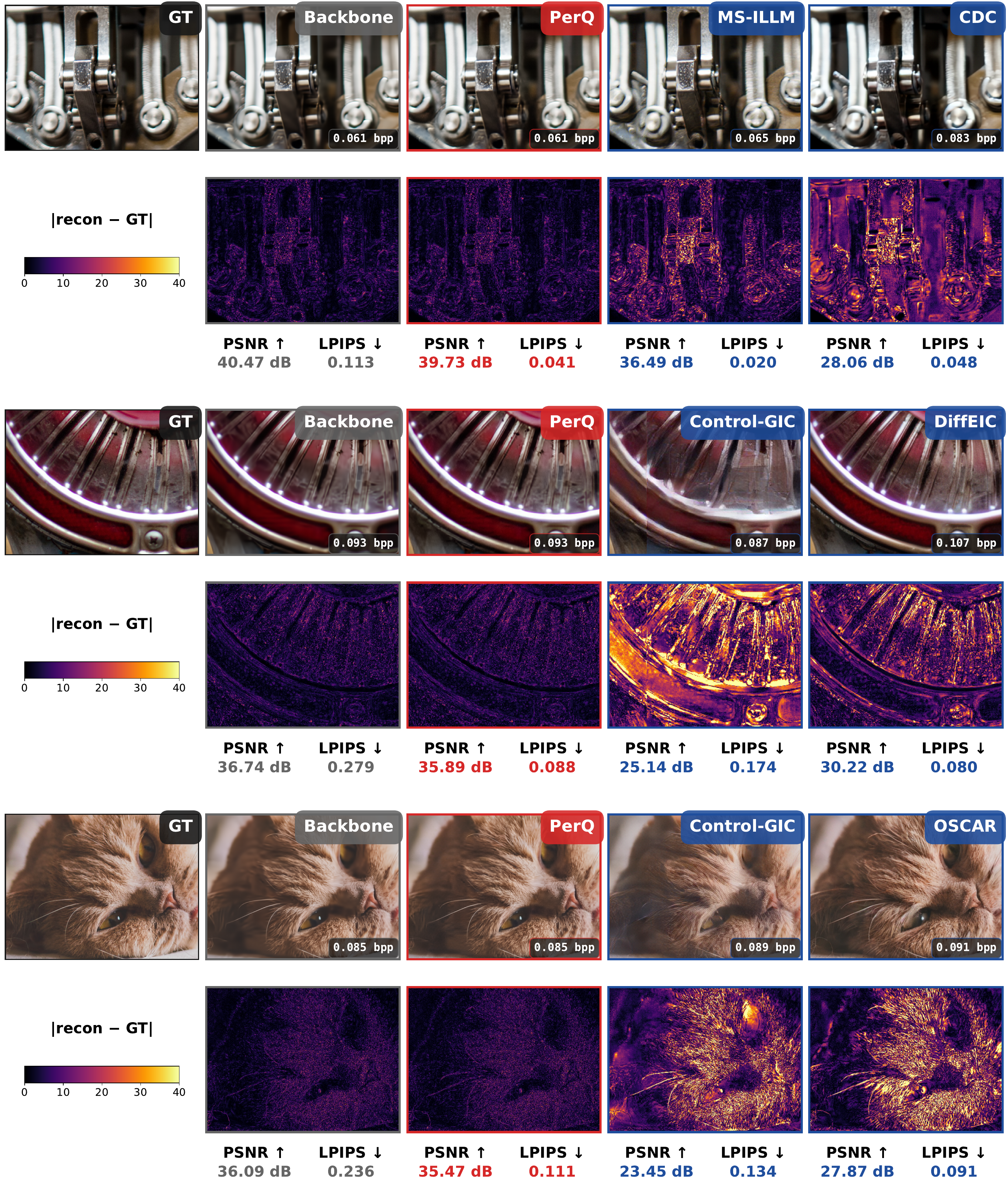}
    \caption{\textbf{Additional visual comparison between PerQ and baseline methods on CLIC2020-test~\cite{toderici2020clic}.}
    Source crop (left) compared with reconstructions from our frozen backbone (\textcolor{gray}{grey}), PerQ
    (\textcolor{red}{red}, ours) at the same rate point, and two generative codecs, GAN-based and diffusion-based
    (\textcolor{blue}{blue}), at similar bitrates. Below each reconstruction, the error map shows its absolute
    difference to the source on a shared color scale, and the numbers give the PSNR and LPIPS of the whole image.}
    \label{fig:more visual}
\end{figure}

\begin{figure}[t]
    \centering
    \includegraphics[width=1\linewidth]{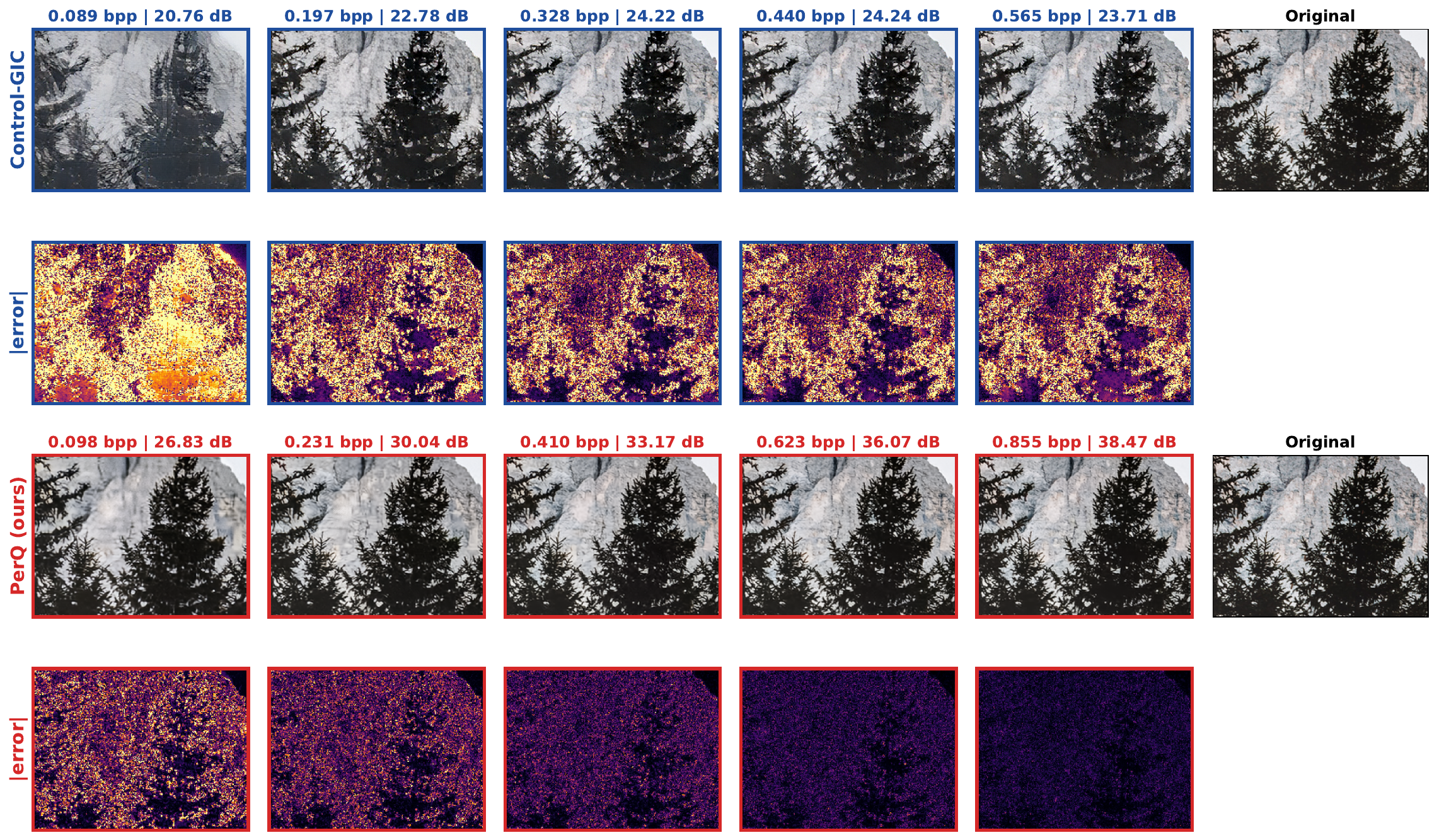}
    \includegraphics[width=1\linewidth]{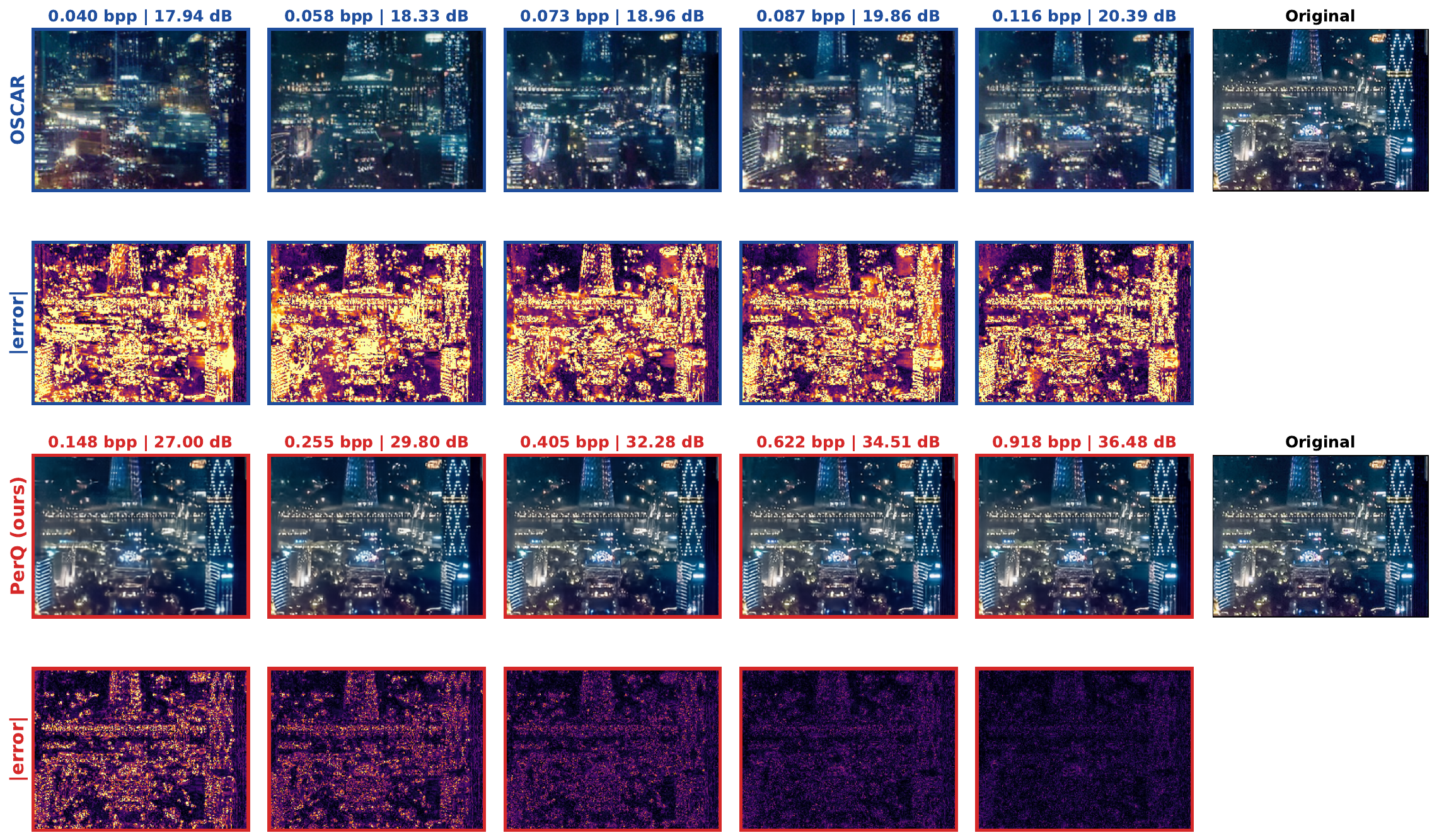}
    \caption{\textbf{Reconstructions across rates on CLIC2020-test~\cite{toderici2020clic}.}
    Generative codecs (\textcolor{blue}{blue}) and PerQ (\textcolor{red}{red}) at their own rate points, with the
    absolute error map below each crop. The PSNR of PerQ keeps rising
    with the rate, while Control-GIC and OSCAR stay below $25$~dB.}
    \label{fig:app_convergence}
\end{figure}

Additional visual comparison results for comparing with baseline methods are provided in Figures~\ref{fig:more visual}, \ref{fig:app_convergence}. 
\ref{fig:more visual} provides more examples for the comparison with the frozen backbone and other generative-based codecs. In all of them, the compensator reduces LPIPS by $53$--$68\%$ (e.g., $0.279 \rightarrow 0.088$ for the lamp) at a cost of only $0.6$--$0.9$~dB in PSNR. The generative codecs at similar bitrates fall $3.2$--$12.0$~dB below PerQ in PSNR, and their error maps show large deviations from the source in textured regions.
\autoref{fig:app_convergence} follows two images across rates. As the rate grows, the error of PerQ fades steadily and its PSNR rises from $26.8$ to $38.5$~dB and from $27.0$ to $36.5$~dB, whereas Control-GIC and OSCAR stay below $25$~dB.

\section{Limitations and Societal Impacts}
\label{app:limitation}

\noindent\textbf{Limitations.} Alongside the positive characteristics, limitations still remain with PerQ. First, it does not reach the highest perceptual quality attained by fully generative codecs, although it remains very competitive on distortion-based metrics and has achieved a better trade-off between fidelity and realism. Moreover, our design relies on a frozen distortion-optimized backbone and jointly fine-tuning the backbone may further improve performance. A unified theoretical framework that jointly optimizes the rate-distortion-perception trade-off should be investigated as future work.

\noindent\textbf{Societal impacts.} This paper proposes an advanced general-purpose learned image codec that can potentially reduce image storage requirements while maintaining both high fidelity and perceptual quality. This capability may also help reduce the carbon footprint of a wide range of related applications.

\clearpage

\end{document}